\documentclass[a4paper,fleqn]{cas-dc}

\usepackage[numbers]{natbib}

\usepackage{amsmath, amssymb, bm}
\usepackage{graphicx, tabularx, multirow, makecell}
\usepackage{enumitem}
\usepackage{hyperref}
\usepackage[capitalise]{cleveref}       

\usepackage{soul}
\usepackage[dvipsnames]{xcolor}
\usepackage{kotex}

\newcommand{\Gi}{\mathcal{G}_\text{int}}
\newcommand{\Ge}{\mathcal{G}_\text{ext}}
\newcommand{\Ti}{\mathcal{T}_\text{int}}
\newcommand{\Te}{\mathcal{T}_\text{ext}}

\def\tsc#1{\csdef{#1}{\textsc{\lowercase{#1}}\xspace}}
\tsc{WGM}
\tsc{QE}
\tsc{EP}
\tsc{PMS}
\tsc{BEC}
\tsc{DE}

\begin{document}
\let\WriteBookmarks\relax
\def\floatpagepagefraction{1}
\def\textpagefraction{.001}
\shorttitle{Neural Graph Simulator for Complex Systems}
\shortauthors{H. Choi et~al.}

\title[mode = title]{Neural Graph Simulator for Complex Systems}

\author[1]{Hoyun Choi}
\credit{Conceptualization, Investigation, Formal analysis, Writing - original draft, Writing - review \& editing}

\affiliation[1]{
    organization={School of Computational Sciences, Korea Institute for Advanced Study},
    addressline={85, Hoegi-ro, Dongdaemun-gu},
    city={Seoul},
    postcode={02455},
    country={Korea}
}

\author[2]{Sungyoep Lee}
\credit{Conceptualization, Writing - review \& editing}
\affiliation[2]{
    organization={Department of Physics and Astronomy, Seoul National University},
    addressline={1, Gwanak-ro, Gwanak-gu},
    city={Seoul},
    postcode={08826},
    country={Korea}
}

\author[3]{B. Kahng}
\credit{Supervision, Writing - review \& editing}
\affiliation[3]{
    organization={CCSS, KI for Grid Modernization, Korea Institute of Energy Technology},
    addressline={21, Kentech-gil},
    city={Naju-si},
    state={Jeollanam-do},
    country={Korea}
}
\cormark[1]
\ead{bkahng@kentech.ac.kr}

\author[4,5]{Junghyo Jo}
\credit{Conceptualization, Supervision, Writing - review \& editing}
\affiliation[4]{
    organization={Department of Physics Education, Seoul National University},
    addressline={1, Gwanak-ro, Gwanak-gu},
    city={Seoul},
    postcode={08826},
    country={Korea}
}
\affiliation[5]{
    organization={Center for Theoretical Physics and Artificial Intelligence Institute, Seoul National University},
    addressline={1, Gwanak-ro, Gwanak-gu},
    city={Seoul},
    postcode={08826},
    country={Korea}
}
\cormark[2]
\ead{jojunghyo@snu.ac.kr}

\cortext[cor1]{Corresponding author}
\cortext[cor2]{Principal corresponding author}

\begin{abstract}
Simulating complex systems governed by coupled ordinary differential equations (ODEs) typically requires explicit governing equations and incurs high computational cost from iterative numerical solvers. We propose the Neural Graph Simulator (NGS), a general-purpose data-driven framework that learns multivariate continuous-time dynamics directly from noisy and incomplete observations, without prior knowledge of the equations. Unlike neural ODE-based approaches that depend on repeated solver evaluations, NGS approximates the state transition integral using elapsed time intervals as direct inputs. This eliminates numerical integration while naturally supporting irregular sampling and heterogeneous time scales. Through its graph network architecture, NGS generalizes to unseen graph structures and larger system sizes, as demonstrated on independently sampled Erdős–Rényi graphs. We evaluate NGS on comprehensive benchmarks covering linear diffusion, nonlinear and chaotic dynamics, systems with stiffness-like behavior, and real-world spatiotemporal datasets. NGS combines competitive accuracy across benchmarks with substantial speedups over numerical solvers in computationally demanding settings. Notably, NGS exhibits strong extrapolation to unseen graph sizes and longer time horizons, and supports non-uniform time steps. These results establish NGS as an efficient and scalable alternative to solver-dependent neural simulators.
\end{abstract}




\maketitle

\section{Introduction}
Simulating dynamics in complex systems with graph structures, ranging from power grids~\cite{witthaut2022collective,lee2024reinforcement} to social networks~\cite{he2020seir, choi2022covid}, typically relies on numerical solvers integrating coupled ordinary differential equations (ODEs).
While recent neural simulators replace explicit physical equations with learnable dynamics, most approaches, including neural ODEs~\cite{chen2018neural} and their graph-based extensions~\cite{poli2019graph,zang2020neural}, retain the same solver-dependent structure, embedding neural networks inside numerical integration loops.

We argue that the core bottleneck of neural simulators is not the expressiveness of the learned dynamics, but the solver loop itself.
Regardless of how accurately the governing equation is approximated, such methods fundamentally require repeated evaluations over intermediate states, leading to high computational cost, sensitivity to stiffness, and rigid coupling to simulation hyperparameters such as step size and error tolerance.

This observation motivates a shift in perspective: instead of learning better instantaneous dynamics and delegating integration to numerical solvers, we directly learn the state transition induced over an arbitrary time interval.
Concretely, we introduce the Neural Graph Simulator (NGS), a solver-free framework that directly approximates the state increment as a function of the current state, elapsed time, and system parameters.

This formulation leads to several direct consequences:
\begin{itemize}
    \item \textbf{Elimination of overhead}, enabling constant inference cost per time step without intermediate evaluations.
    \item \textbf{Native support for irregular time intervals}, as NGS explicitly conditions the learned state transition.
    \item \textbf{Robustness to rapidly evolving dynamics}, where numerical solvers typically require prohibitively small step sizes.
    \item \textbf{Generalization across graph structures and sizes}, achieved through local, parameter-shared transition rules.
\end{itemize}

We rigorously evaluate the NGS on a comprehensive benchmark suite spanning linear to nonlinear chaotic dynamics, static to temporal graphs with evolving topology, and synthetic to real-world datasets.
NGS matches or exceeds the accuracy of the evaluated neural baselines in most synthetic benchmark settings and remains competitive with specialized models on real-world forecasting tasks.
Its computational advantages are particularly pronounced on larger graphs and in temporal-graph simulations, where it substantially reduces runtime relative to conventional numerical solvers.

\section{Related Work} \label{sec:related}

\paragraph{Continuous-time graph dynamics.}
Integrating Graph neural networks (GNNs)~\cite{gori2005new, scarselli2008graph} with neural ordinary differential equations (neural ODEs) has significantly advanced continuous-time dynamics modeling.
Seminal frameworks such as GDE~\cite{poli2019graph} and NDCN~\cite{zang2020neural} combined continuous-depth GNNs with numerical solvers, establishing benchmarks in domains like multi-agent systems and traffic forecasting.
Subsequent studies extended this foundation: HOPE~\cite{luo2023hope} utilizes second-order neural ODEs for long-term dependencies, Graph NeuralSDE~\cite{bergna2023graph} incorporates stochasticity, and SGODE~\cite{chen2024signed} addresses signed interaction graphs.
These methods are compelling when the goal is to learn smooth dynamics; however, their inference cost is intrinsically tied to the inner numerical integration, which becomes a bottleneck during long rollouts or in chaotic or stiff regimes.
In contrast to recurrent models that use $\Delta t$ to regulate hidden state evolution via exponential decay or attention weights~\cite{che2018recurrent,tan2020data}, NGS learns a $\Delta t$-conditioned state transition integral in a single forward pass.
This physically grounded role eliminates the inner Euler/Runge--Kutta loop while preserving continuous-time queries at varying $\Delta t$ within the range represented during training.

\paragraph{Learned simulators.}
Graph-based neural simulators such as GNS~\cite{sanchez2020learning} demonstrate that local interaction models can generalize across long horizons and substantially larger systems by leveraging shared message-passing updates.
This perspective aligns with NGS: both treat dynamics as computations over local neighborhoods and rely on parameter sharing to transfer across system sizes.
However, most learned simulators are fundamentally discrete-time (fixed-step) predictors, and adapting them to heterogeneous $\Delta t$ typically requires ad hoc reparameterization.
NGS elevates $\Delta t$ to a first-class input and directly models transitions over non-uniform intervals, which is crucial for scientific data where sampling rates change or measurements are missing.

\paragraph{Learning with corrupted data.}
Robustness to partially observed data is crucial for real-world deployment.
Latent-variable continuous-time models~\cite{huang2020learning} can infer trajectories from sparse samples, but they may degrade when certain nodes remain persistently unobserved.
Recent methods like GPT-ST~\cite{li2023generative} (masked autoencoding) and OPCR~\cite{deng2024learning} (confidence-based refinement) address this with sophisticated imputation strategies.
Instead, NGS adopts a lightweight structural masking strategy tailored to local simulators (described in \cref{subsec:NGS}), rather than explicit imputation.
This is particularly compatible with NGS’s strict parameter sharing and local-update structure.

\paragraph{Size generalization in GNNs.}
Although theoretically size-invariant, GNNs can lead to suboptimal solutions when generalizing from small to large graphs due to local structural discrepancies~\cite{yehudai2021local}.
Strategies such as SizeShiftReg~\cite{buffelli2022size} and DISGEN~\cite{huang2024disgen} improve robustness via regularization or representation disentanglement, which disentangles size factors from representations.
Such advances have shown promise in molecular science~\cite{wang2021scalable}, infrastructure systems~\cite{nauck2022predicting, jhun2023prediction}, and biology~\cite{li2024size}.
NGS achieves scale invariance through its local $L$-hop ego-graph prediction mechanism with shared parameters, making the per-node transition rule independent of global graph size.

\paragraph{Temporal and evolving graphs.}
Temporal graph structures driven by either exogenous factors or node states present unique challenges for standard neural ODEs.
Prominent solutions include TDE-GNN~\cite{eliasof2024temporal}, which uses multi-head attention for higher-order derivatives, and TG-ODE~\cite{gravina2024tgode}, which employs recursive TAGConv~\cite{du2018topology} for varying topologies and irregular time steps.
Additionally, EGODE~\cite{yuan2024egode} introduces hierarchical graph structures to model abrupt topological changes in rigid systems.
NGS supports temporal graphs by treating the evolving structure as a sequence of graph snapshots and applying the same Markovian transition rule at each step.

\section{Methodology}   \label{sec:method}
\begin{figure*}
    \centering
    \includegraphics[width=\linewidth]{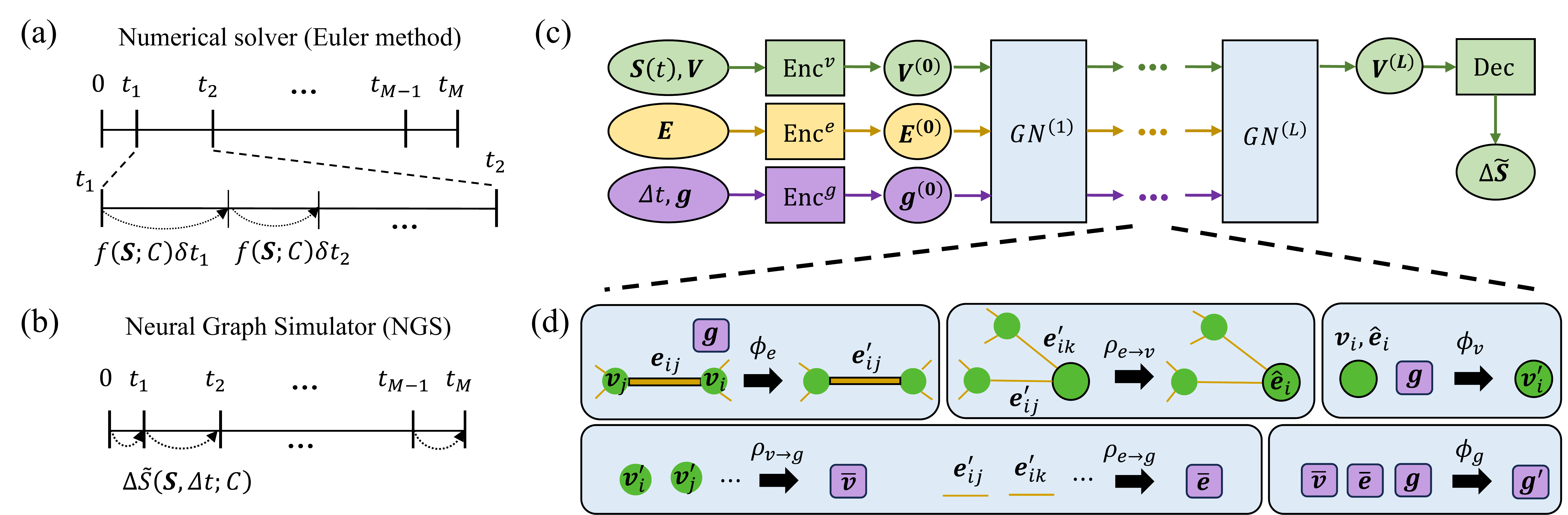}
    \caption{
        \textbf{Overview of the Neural Graph Simulator.}
        (a) Traditional numerical solvers integrate dynamics through iterative sub-steps of $\delta t$.
        (b) NGS directly approximates the integral $\Delta \tilde{\bm{S}}$ for an arbitrary interval $\Delta t$ in a single forward pass.
        (c) Overview of the NGS architecture. Inputs---state $\bm{S}(t)$, time step $\Delta t$, and parameters $C=(\bm{V}, \bm{E}, \bm{g})$—are encoded, processed via $L$ Graph Network (GN) layers, and decoded to update the state.
        (d) Detailed computation flow within a single GN layer, involving update ($\phi$) and aggregation ($\rho$) functions.
    }   \label{fig:schematic}
\end{figure*}

\subsection{Preliminaries}  \label{subsec:preliminaries}
We consider an autonomous system governed by coupled ODEs:
\begin{equation}    \label{eq:governing}
    \frac{d\bm{s}_i(t)}{dt} = f(\bm{S}; C), \ {}^\forall i \in \{1, 2, \dots N\}.
\end{equation}
The homogeneous node dynamics, in which each of the $N$ nodes evolves according to \cref{eq:governing}, allow parameter sharing across nodes, facilitating trajectory inference even with missing values.
Here, $\bm{S} \equiv [\bm{s}_1, \dots, \bm{s}_N]$ represents collective node states, with $\bm{s}_i \in \mathbb{R}^d$.
The system parameter $C$ contains all known, possibly time-varying exogenous inputs that modulate the dynamics.
Specifically, $C$ encompasses node coefficients $\bm{V} \equiv [\bm{v}_1, \dots, \bm{v}_N]$, edge coefficients $\bm{E} \equiv [\bm{e}_1, \dots, \bm{e}_E]$, and a global coefficient $\bm{g}$.
Given an initial condition $\bm{S}(0)$, a trajectory is defined as $\{\bm{S}(t_m)\}_{m=1}^M$ over the time domain $[0, T]$ with non-uniform intervals $\Delta t_m \equiv t_{m+1} - t_m$.
Note that \cref{eq:governing} defines dynamics on a discrete graph structure, distinct from PDEs posed over continuous Euclidean domains.

Conventional numerical solvers~\cite{dormand1980family, hairer2008solving, petzold1983automatic} require fine-grained steps $\delta t \ll \Delta t$ to solve \cref{eq:governing}, necessitating iterative sub-steps that incur significant computational overhead, as depicted in \cref{fig:schematic}(a).

\subsection{Neural Graph Simulator} \label{subsec:NGS}
We target the exact state increment over an arbitrary interval $\Delta t$,
\begin{equation} \label{eq:ngs}
    \Delta \bm{S}(\bm{S}(t), \Delta t; C) \equiv \int_{t}^{t+\Delta t} f(\bm{S}(\tau); C) d\tau,
\end{equation}
which is determined by the current state $\bm{S}(t)$, the time interval $\Delta t$, and the system parameters $C$ under the autonomous dynamics in \cref{eq:governing}.
Instead of approximating this integral via repeated solver steps (as in NDCN or GDE), NGS learns a direct estimator $\Delta \tilde S_{\theta}(S(t),\Delta t;C)$ in a single forward pass.

Importantly, \cref{eq:ngs} imposes several structural constraints on any practical estimator.
First, the integral does not require access to intermediate states $S(\tau)$, precluding recursive evaluation.
Second, the elapsed time $\Delta t$ must explicitly condition the estimator, as the same initial state may induce qualitatively different transitions over different time horizons.
Third, to ensure scalability and generalization across graph sizes, the estimator must operate locally with shared parameters, without introducing dependence on global graph cardinality.

The NGS (\cref{fig:schematic}(c)) processes three input modalities: current state $\bm{S}(t)$, time step $\Delta t$, and parameters $C=\{\bm{V}, \bm{E}, \bm{g} \}$.
Encoders first map these inputs to high-dimensional latent vectors.
Specifically, encoded $\bm{S}(t)$ is concatenated with node features, while encoded $\Delta t$ is appended to global features.
Consistent with GNN principles, all encoders share weights across nodes and edges through identical neural networks.
A key design choice is to encode $\Delta t$ and system-level parameters as global features.
Naively duplicating these scalars across all nodes or edges, as commonly done in standard message-passing neural networks (MPNNs)~\cite{gilmer2017neural}, implicitly couples the learned transition to graph size and density.
In contrast, the global channel in Graph Network (GN) layers~\cite{battaglia2018relational} allows such quantities to influence node and edge updates through shared aggregation, allowing the same learned update functions to be applied across graphs of different sizes.

Latent representations are updated through $L$ GN layers (\cref{fig:schematic}(d)).
The GN layer, which generalizes MPNNs, updates node, edge, and global features using trainable update functions and non-trainable permutation-invariant aggregation functions.
We use $L=2$ across all experiments as it provides a strong accuracy-efficiency tradeoff.

Finally, a parameter-sharing decoder transforms the processed node features $\bm{V}^{(L)}$ into state increments $\Delta \tilde{\bm{S}}$, yielding the next state $\tilde{\bm{S}}(t + \Delta t) = \bm{S}(t) + \Delta \tilde{\bm{S}}$.
See Appendix~\ref{app:NGS} for architectural details.

\paragraph{Graph generalization.}
NGS applies shared node and edge update functions to local neighborhoods, allowing the same trained model to process graphs with different connectivity patterns and sizes.
Our static-graph experiments demonstrate generalization to unseen Erdős–Rényi (ER) graph realizations, including graphs substantially larger than those used for training.
Empirically, we find that small values of $L$ are sufficient to capture the effective propagation of dynamics across all considered systems.

\paragraph{Temporal generalization.}
Conditioning on non-uniform $\Delta t$ allows NGS to predict transitions over non-uniform intervals and results in continuous-time solutions.
Autoregressive prediction further allows simulation beyond the training horizon, provided the system state remains within the valid domain of the learned dynamics.

\begin{figure}
    \centering
    \includegraphics[width=\linewidth]{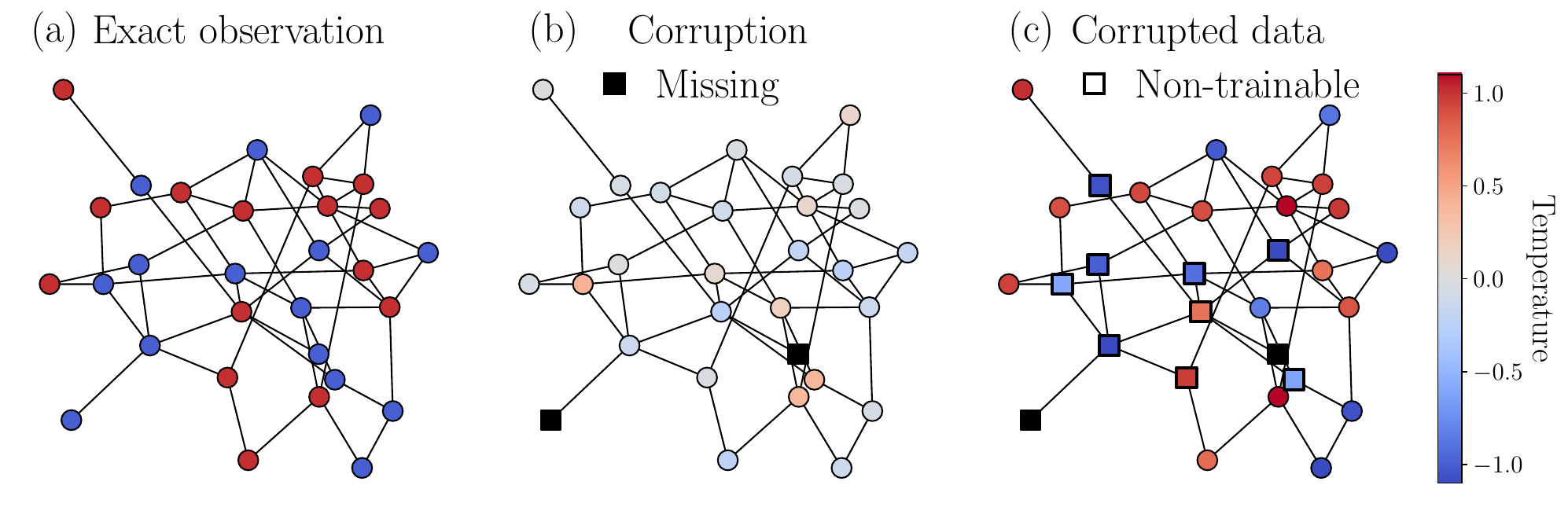}
    \caption{
        \textbf{Data corruption in heat diffusion.}
        (a) Ground truth state of a heat diffusion system (color indicates temperature).
        (b) Data corruption visualization: Gaussian noise (color distortion) and missing nodes (black squares).
        (c) NGS training strategy with $L=2$.
        To prevent missing value propagation, gradients are masked for missing nodes and their $L$-hop neighbors (squares), ensuring learning occurs only on valid subgraphs.
    }   \label{fig:corrupted}
\end{figure}

\paragraph{Learning from corrupted data.}
To encompass practical scenarios, we implement Gaussian noise with standard deviation $\sigma$ and random node masking of fraction $p$.
\cref{fig:corrupted}(a) and (b) demonstrate this process in a heat diffusion system.
Crucially, NGS employs a structural masking strategy: gradients from missing nodes and their $L$-hop neighbors are zeroed out during training (\cref{fig:corrupted}(c)).
This ensures the model learns solely from valid, causally intact subgraphs, recovering true dynamics despite data corruption.

\subsection{Training and Evaluation Protocols}  \label{subsec:train_eval}
NGS is trained via single-step prediction, minimizing the mean squared error (MSE) between predicted states $\bm{S}(t) + \Delta \tilde{\bm{S}}$ and ground truth $\bm{S}(t+\Delta t)$ across all trainable nodes and temporal points.
This purely data-driven approach requires no prior knowledge of the governing equations.

For inference, NGS performs autoregressive rollout as illustrated in \cref{fig:schematic}(b).
The simulation starts from the initial condition $\tilde{\bm{S}}(t_0) \equiv \bm{S}(t_0)$ and recursively predicts subsequent states $\tilde{\bm{S}}(t_{m+1})$ based on the prior estimate $\tilde{\bm{S}}(t_m)$.

\begin{figure*}
    \centering
    \includegraphics[width=\linewidth]{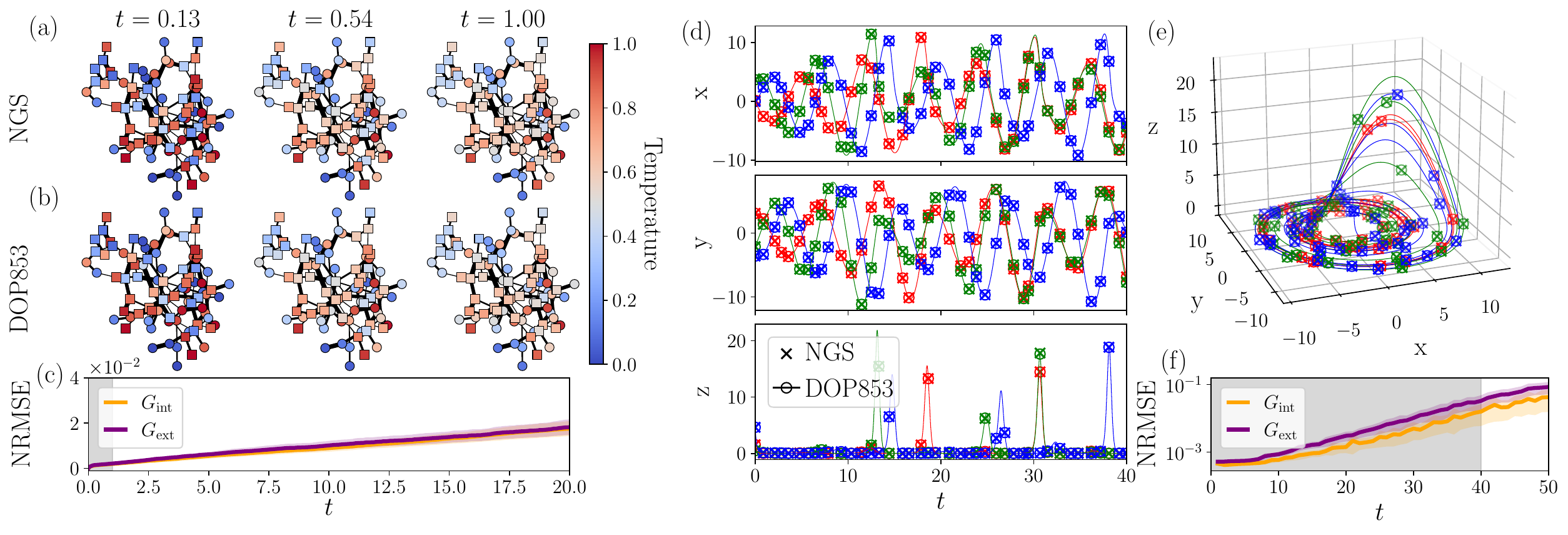}
    \caption{
        \textbf{Simulation snapshots of heat diffusion and coupled Rössler systems.}
        (a, b) Snapshots of the heat diffusion system simulated by NGS and the reference solver (DOP853).
        Node color and edge thickness indicate temperature and dissipation rate, respectively.
        Squares represent nodes with their gradient masked during training.
        (c) Time evolution of NRMSE between NGS and DOP853 trajectories for $\Gi$ (orange) and $\Ge$ (purple).
        Solid curves show the mean NRMSE within each graph domain, and shaded bands indicate pointwise 95\% confidence intervals for the means.
        As $\Gi$ and $\Ge$ curves nearly overlap, they appear as a single curve.
        Gray shading marks the training interval $\Ti$.
        (d, e) Trajectories of $x$, $y$, and $z$ coordinates for three random nodes (red, green, blue) in the coupled Rössler system, demonstrating NGS's ability to capture chaotic attractors.
        (f) Corresponding NRMSE for the coupled Rössler system, with the same conventions as in (c).
    }   \label{fig:static}
\end{figure*}

\paragraph{Accuracy metrics.}
We quantify prediction accuracy using the normalized root mean squared error (NRMSE):
\begin{equation}    \label{eq:nrmse}
    \text{NRMSE} = \frac{1}{std(\bm{S})} \sqrt{ \frac{1}{MNd} \sum_{m=1}^{M} \sum_{i=1}^N \lVert \bm{s}_i(t_m) - \tilde{\bm{s}}_i(t_m) \rVert_2^2},
\end{equation}
where $\lVert \cdot \rVert_2$ denotes the $\ell_2$ norm and $std(\bm{S})$ is the standard deviation of the true trajectory.
This dimensionless metric enables meaningful cross-system comparisons by normalizing prediction errors against inherent system variability.

\paragraph{Efficiency metrics.}
Computational efficiency is measured through two complementary metrics: the number of function evaluations (NFE) and wall-clock runtime.
NFE counts total calls to the governing equation (numerical solvers) or neural network forward passes (deep learning approaches).
Runtime measurements provide additional insights, capturing the aggregated computational cost, including overheads from complex nonlinear operations.
Each experimental setting is designed to isolate and validate a distinct advantage of the proposed formulation.

\section{Experiments}   \label{sec:experiment}

\begin{table*}[ht]
    \renewcommand{\arraystretch}{1.2}
    \centering
    \begin{tabular}{l*{4}{c}lcc}
        \toprule
        \multirow{2}{*}{Model} & \multicolumn{4}{c}{NRMSE ($\times 10^{-3}$)} &  & \multicolumn{2}{c}{NFE / Runtime (sec)} \\
        \cline{2-5} \cline{7-8}
        & ($\Gi$, $\Ti$) & ($\Gi$, $\Te$) & ($\Ge$, $\Ti$) & ($\Ge$, $\Te$) &  & $\Gi$ & $\Ge$ \\
        \midrule
        DOP853 & - & - & - & - &  & 1051 / 0.01 & 1201 / 3.92 \\
        \midrule
        $\text{NDCN}_\text{euler}$ & $3.0 \pm 0.5$ & $17.3 \pm 3.8$  & $3.1 \pm 0.4$ & $12.6 \pm 2.2$ &  & 400 / 0.07  & 400 / 0.15  \\
        $\text{NDCN}_\text{dopri5}$ & $3.1 \pm 0.5$ & $24.0 \pm 1.6$ & $3.3 \pm 0.4$ & $19.3 \pm 1.2$ &  & 815 / 0.15  & 819 / 0.30  \\
        $\text{GDE}_\text{euler}$ & $3.1 \pm 0.9$ & $15.7 \pm 3.2$ & $3.2 \pm 1.1$ & $10.9 \pm 4.0$ &  & 400 / 0.20  & 400 / 0.35  \\
        $\text{GDE}_\text{dopri5}$ & $2.6 \pm 0.9$ & $\bm{11.8 \pm 4.8}$ & $2.7 \pm 0.7$ & $\bm{9.9 \pm 2.5}$ &  & 1265 / 0.65 & 1411 / 1.22 \\
        \midrule
        NGS (ours) & $\bm{1.4 \pm 0.1}$ & $12.1 \pm 4.2$ & $\bm{1.5 \pm 0.2}$ & $\bm{9.9 \pm 1.5}$ &  & 400 / 0.15 & 400 / 0.15 \\
        \bottomrule
    \end{tabular}
    \caption {
        \textbf{Performance on linear heat diffusion system.}
        Comparison of normalized root mean squared error (NRMSE) and efficiency metrics across diverse domains.
        Subscripts `euler' and `dopri5' indicate the ODE solver backend.
        NRMSEs are reported with 95\% confidence intervals in units of $10^{-3}$ and NFE and runtime are averaged over $\Ti \cup \Te$.
    } \label{tab:heat}
\end{table*}

\begin{table*}[ht]
    \renewcommand{\arraystretch}{1.2}
    \centering
    \begin{tabular}{l*{4}{c}lclcc}
        \toprule
        \multirow{2}{*}{Model} & \multicolumn{4}{c}{NRMSE ($\times 10^{-2}$)} &  & $\lambda_{0.01}$ &  & \multicolumn{2}{c}{NFE / Runtime (sec)} \\
        \cline{2-5} \cline{7-7} \cline{9-10}
        & ($\Gi$, $\Ti$) & ($\Gi$, $\Te$) & ($\Ge$, $\Ti$) & ($\Ge$, $\Te$) &  & $\Gi \cup \Ge$ &  & $\Gi$ & $\Ge$ \\
        \midrule
        DOP853 & - & - & - & - &  & 0.025 &  & 11970 / 0.14 & 13115 / 31.9 \\
        \midrule
        $\text{NDCN}_\text{euler}$ & $8.5 \pm 1.2$ & $53.1 \pm 1.0$ & $9.0 \pm 1.1$ & $52.6 \pm 0.8$ & & 0.007 &  & 50 / 0.01 & 50 / 0.02 \\
        $\text{NDCN}_\text{dopri5}$ & $7.0 \pm 1.8$ & $49.8 \pm 2.3$ & $7.2 \pm 1.8$ & $48.3 \pm 1.8$ & & 0.004 &  & 138 / 0.03 & 138 / 0.05 \\
        $\text{GDE}_\text{euler}$ & $9.2 \pm 1.2$ & $52.5 \pm 4.3$ & $9.6 \pm 1.1$ & $54.2 \pm 4.1$ & & 0.004 &  & 50 / 0.03 & 50 / 0.04 \\
        $\text{GDE}_\text{dopri5}$ & $9.1 \pm 1.2$ & $53.3 \pm 0.9$ & $8.9 \pm 1.1$ & $52.7 \pm 0.8$ & & 0.004 &  & 129 / 0.08 & 129 / 0.15 \\
        \midrule
        NGS (ours) & $\bm{2.3 \pm 1.1}$ & $\bm{8.6 \pm 4.0}$ & $\bm{3.2 \pm 0.8}$ & $\bm{13.8 \pm 3.7}$ &  & 0.026 &  & 50 / 0.02 & 50 / 0.03 \\
        \bottomrule
    \end{tabular}
    \caption {
        \textbf{Performance on nonlinear, chaotic coupled Rössler system.}
        Comparison of NRMSE in units of $10^{-2}$ and Lyapunov exponent $\lambda_{0.01}$.
        The latter is obtained using Gaussian initial perturbations with a standard deviation of $0.01$ and averaged over $\Gi \cup \Ge$.
        Its proximity to $\lambda_{0.01} \approx 0.025$ of the ground truth indicates simulation fidelity.
    } \label{tab:rossler}
\end{table*}

We emphasize that the ``mathematical systems'' benchmarks represent canonical scientific workloads where governing dynamics are known and solver cost is a primary bottleneck.
These settings also allow controlled evaluation of fidelity and efficiency under various noise levels, missingness, and heterogeneous sampling, complemented by real-world spatiotemporal datasets.

\subsection{Mathematical Systems on Static Graphs}  \label{subsec:static}
We first validate NGS on two canonical systems: heat diffusion (linear) and coupled Rössler attractors (nonlinear and chaotic).

\paragraph{Heat diffusion system.}
Governed by linear thermal interactions:
\begin{equation}    \label{eq:heat}
    \frac{dT_i}{dt} = \sum_{j \in \mathcal{N}(i)} d_{ij} (T_j - T_i),
\end{equation}
where $\mathcal{N}(i)$ denotes the set of its neighbors.
This system represents one of the few complex dynamical systems with analytical solutions, serving as a reliable benchmark for accuracy verification.
Nodes possess temperature $T_i$ and interact via edges with dissipation rate $d_{ij}$.

\paragraph{Coupled Rössler system.}
To test NGS's ability to model nonlinear chaotic dynamics, we employ coupled Rössler attractors:
\begin{equation}    \label{eq:rossler}
    \begin{split}
        \frac{dx_i}{dt} & = -y_i -z_i                                                   \\
        \frac{dy_i}{dt} & = x_i + ay_i + \sum_{j \in \mathcal{N}(i)} K_{ij} (y_j - y_i) \\
        \frac{dz_i}{dt} & = b + z_i (x_i - c)
    \end{split}
\end{equation}
Each node $i$ represents an attractor in  $\mathbb{R}^3$, coupled along the $y$-axis with strength $K_{ij}$.
This system exhibits chaotic oscillations and sudden jumps along the $z$-axis, posing significant challenges for long-term prediction.
This experiment tests the stability of the NGS under exponential sensitivity to initial conditions.

\paragraph{Protocol.}
We generate ground truth trajectories using the adaptive 8th-order Runge--Kutta solver (DOP853)~\cite{hairer2008solving}.
To evaluate generalization to unseen structure and larger graph sizes, we construct two independently sampled ER graph ensembles: an interpolation set $\Gi$ with $N \in [100, 200]$, and an extrapolation set $\Ge$ with $N \in [2000, 3000]$.
Similarly, time domains are split into interpolation $\Ti \equiv [0, T_\text{int}]$ and extrapolation $\Te \equiv [T_\text{int}, T_\text{ext}]$.
A single NGS model is trained on the corrupted small-scale domain ($\Gi, \Ti$) and evaluated across all subsequent evaluations without fine-tuning, regardless of graph sizes and time horizons.
Crucially, training data is corrupted with Gaussian noise ($\sigma=0.001$) and missing nodes ($p=0.1$).
Detailed experiment configurations and results for varying $\sigma$ and $p$ are presented in Appendix~\ref{app:mathematical}.

\subsubsection{Reconstruction of corrupted data}   \label{subsubsec:reconstruction}
We first assess the model's ability to recover true dynamics from trajectories in the training data.
Nodes designated as missing during training have their initial states revealed at inference, and the model performs forward autoregressive rollout to predict the full trajectory of the entire system.

\paragraph{Heat diffusion system.}
\cref{fig:static}(a, b) visualize the thermal evolution simulated by NGS versus the ground truth (DOP853).
Despite training on noisy, incomplete data, NGS accurately reconstructs the temperature field.
Quantitatively, NGS achieves a reconstruction NRMSE of $(2.24 \pm 0.05) \times 10^{-3}$.
Throughout the experiments, we report means with 95\% normal-approximation confidence intervals based on sampled standard errors across 50 independent systems, with a fixed trained model.
This performance surpasses the noise level of the training data ($\sigma / std(\bm{S}) \approx 4.06 \times 10^{-3}$), indicating that NGS learns the underlying physical laws from corrupted data.

\paragraph{Coupled Rössler system.}
In the more challenging coupled Rössler system, NGS maintains robustness.
\cref{fig:static}(d, e) show that the model captures characteristic attractor behaviors, exhibiting chaotic $xy$-plane oscillations with occasional synchronization between red and green nodes during $t \in [25, 35]$ and abrupt $z$-axis spikes.
It achieves an NRMSE of $(1.88 \pm 0.17) \times 10^{-2}$, successfully tracking dynamics at non-uniform time points.

\subsubsection{Generalization over graph and time}   \label{subsubsec:nonstiff_generalization}
We extend the evaluation to four domain combinations: ($\Gi, \Ti$), ($\Gi, \Te$), ($\Ge, \Ti$), and ($\Ge, \Te$).
We set $(T_\text{int}, T_\text{ext})=(1,20)$ for heat diffusion and $(40,50)$ for the coupled Rössler system.
Tables \ref{tab:heat} and \ref{tab:rossler} benchmark NGS against state-of-the-art neural ODE variants (NDCN~\cite{zang2020neural}, GDE~\cite{poli2019graph}) across varying graph sizes and time horizons.
As these baselines provide no mechanism to handle missing nodes, they were trained exclusively on complete data ($p=0$).

\begin{figure*}
    \centering
    \includegraphics[width=\linewidth]{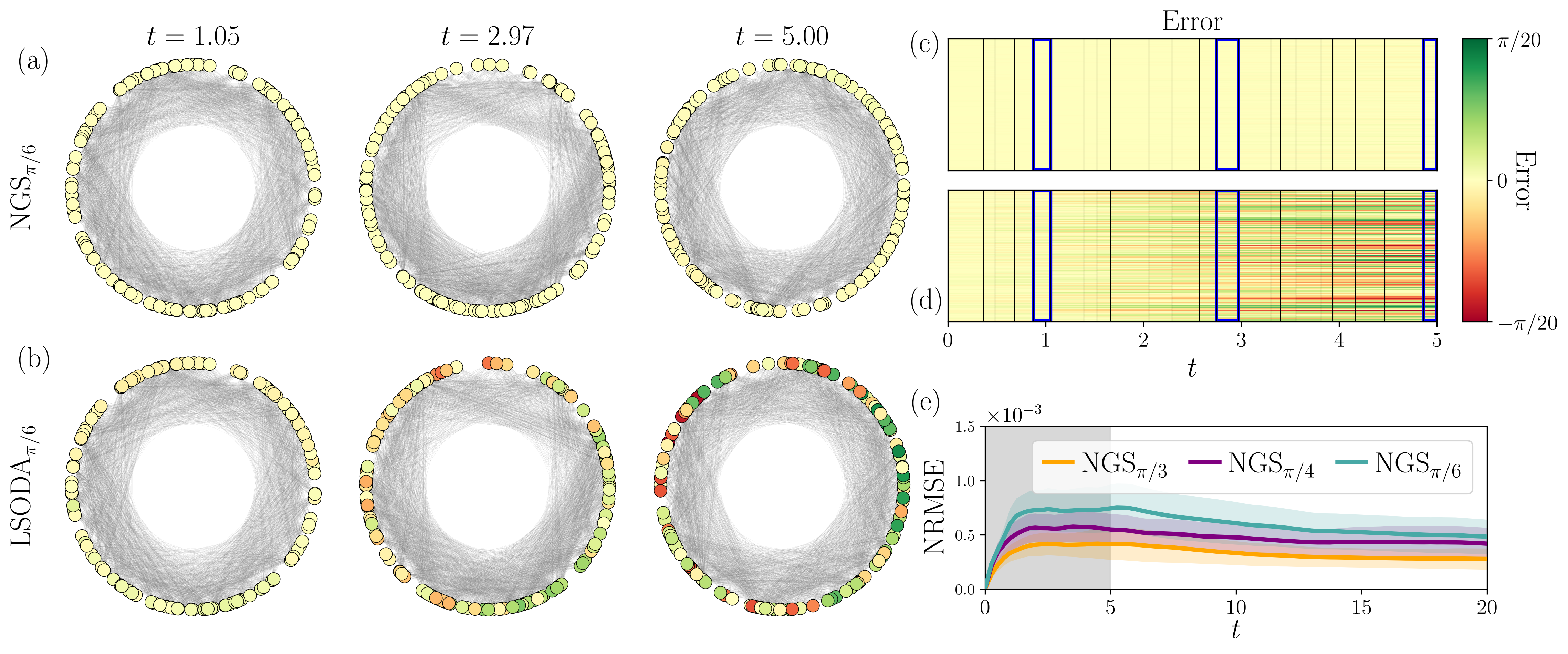}
    \caption{
        \textbf{Simulation snapshots and error evolution of the Kuramoto system.}
        (a), (b) Snapshots of the Kuramoto system simulated by NGS and LSODA with $\theta_\text{th}=\pi/6$, where edges satisfy $| \sin (\theta_j-\theta_i-\pi/2 ) | \le \sin(\pi/6)=1/2$.
        Node positions indicate their phase simulated by each solver, with colors showing errors relative to ground truth ($\theta_\text{th}=\pi/2$).
        (c), (d) Errors of all nodes over evaluation time, utilizing the same color scheme, with black lines indicating non-uniform evaluation times and blue squares highlighting the snapshots presented in (a) and (b).
        (e) Time evolution of NRMSE for $\theta_\text{th}=\pi/3$, $\pi/4$, and $\pi/6$.
        Solid curves show the average NRMSE over 50 test systems, and shaded bands represent 95\% confidence intervals.
        The gray background marks the training horizon $\Ti$: $0 \le t \le 5$.
    }   \label{fig:temporal}
\end{figure*}

\paragraph{Accuracy.}
NGS maintains comparable predictive performance across $\Gi$ and $\Ge$ within each temporal domain without fine-tuning.
While it achieves competitive accuracy on linear heat diffusion, its advantage over the neural ODE baselines is more pronounced on the nonlinear coupled Rössler system, especially in the temporal extrapolation domain $\Te$.
Figs.~\ref{fig:static}(c) and (f) show that autoregressive rollout errors grow approximately linearly in the former and exponentially in the latter across both graph domains.
This exponential growth is consistent with the chaotic Rössler system's sensitivity to initial conditions.
As a complementary measure of dynamical fidelity, we compare maximal Lyapunov exponent estimates $\lambda_{0.01}$, computed using Gaussian initial perturbations with a standard deviation of $0.01$ and averaged over $\Gi \cup \Ge$.
As shown in \cref{tab:rossler}, NGS yields $\lambda_{0.01} \approx 0.026$, closely matching the reference value of $0.025$ despite increasing trajectory errors.
In contrast, the neural ODE baselines substantially underestimate this value, yielding only  $0.004$--$0.007$.

\paragraph{Efficiency.}
While NGS is comparable to Euler-based neural ODEs, it achieves speedups over high-order solvers (dopri5) and maintains constant inference cost even as graph size scales ($\Ge$), whereas traditional solvers suffer from exploding computational overhead (\cref{tab:heat} and \cref{tab:rossler}).
In addition, NGS exhibits linear memory scaling $O(N + E)$ and constant inference cost per step across graph sizes.

\begin{table*}
    \renewcommand{\arraystretch}{1.2}
    \centering
    \begin{tabular}{lccclccc}
        \toprule
        \multirow{2}{*}{Model} & \multicolumn{3}{c}{NRMSE ($\times 10^{-3}$)} & & \multicolumn{3}{c}{NFE / Runtime (sec)} \\
        \cline{2-4} \cline{6-8}
        & $\theta_\text{th}=\pi/3$ & $\theta_\text{th}=\pi/4$ & $\theta_\text{th}=\pi/6$ &  & $\theta_\text{th}=\pi/3$ & $\theta_\text{th}=\pi/4$ & $\theta_\text{th}=\pi/6$ \\
        \midrule
        LSODA & $3.79 \pm 0.34$ & $11.43 \pm 1.17$ & $22.64 \pm 2.43$ &  & $2 \times 10^8$ / 6228 & $4 \times 10^8$ / 12190 & $1 \times 10^9$ / 13851 \\
        \midrule
        $\text{TG-ODE}$ & $3.26 \pm 0.14$ & $4.81 \pm 0.23$ & $5.46 \pm 0.16$ &  & 80 / 0.417 & 80 / 0.406 & 80 / 0.371 \\
        $\text{TDE-GNN}$ & $2.10 \pm 0.13$ & $2.86 \pm 0.16$ & $3.28 \pm 0.20$ &  & 80 / 0.495 & 80 / 0.430 & 80 / 0.419 \\
        \midrule
        NGS (ours) & $\bm{0.30 \pm 0.10}$ & $\bm{0.45 \pm 0.14}$ & $\bm{0.55 \pm 0.17}$ &  & 80 / 0.056 & 80 / 0.054 & 80 / 0.054 \\
        \bottomrule
    \end{tabular}
    \caption {
        \textbf{Comparison on the Kuramoto system with stiffness-like behavior.}
        Performance under varying interaction thresholds $\theta_\text{th}$, where lower $\theta_\text{th}$ induces sparser graphs.
        NGS maintains high accuracy and constant efficiency, whereas the adaptive solver (LSODA) suffers from exploding computational costs measured over $\Ti \cup \Te$.
    } \label{tab:kuramoto}
\end{table*}

\subsection{Mathematical Systems on Temporal Graphs}  \label{subsec:temporal}
The Kuramoto model describes phase synchronization among coupled oscillators:
\begin{equation}    \label{eq:kuramoto}
    \frac{d \theta_i}{dt} = \omega_i + \frac{K}{N}\sum_{j \neq i} \sin (\theta_j - \theta_i),
\end{equation}
where $\theta_i \in (-\pi, \pi]$ is the phase, $\omega_i$ the intrinsic angular velocity, and $K$ the coupling strength.
Although the classical model assumes complete connectivity between oscillators, implementing $N^2$ interactions becomes computationally intractable for large systems.

To mitigate this, we employ a threshold-based sparse approximation: interactions are active only when $|\sin(\theta_j - \theta_i)|$ exceeds a certain threshold.
Specifically, for $0<\theta_\text{th} \le \pi/2$, interactions are considered active when $| \sin (\theta_j-\theta_i-\pi/2 ) | \le \sin \theta_\text{th}$.
At $\theta_\text{th}=\pi/2$, all pairwise interactions are retained, corresponding to the static, fully connected graph.
This creates a state-dependent temporal graph, where edges switch as node states evolve.
This setting allows us to evaluate NGS on temporal graphs.
Topology switching also introduces discontinuities in the right-hand side of the thresholded Kuramoto equations.
DOP853 was prohibitively slow in our initial simulations, exhibiting stiffness-like behavior.

\paragraph{Baselines and protocol.}
Ground truth trajectories are generated from the fully connected Kuramoto system ($\theta_\text{th}=\pi/2$) using DOP853.
We use LSODA~\cite{petzold1983automatic} as the numerical baseline to integrate the thresholded Kuramoto equations.
For neural baselines, we compare against well-established temporal graph models: TDE-GNN~\cite{eliasof2024temporal} and TG-ODE~\cite{gravina2024tgode}.
NGS and the neural baselines are trained on the ground truth trajectories using thresholded input graphs.
The training horizon is $T_\text{int}=5$, and evaluation extends to $T_\text{ext}=20$.
Evaluation focuses on varying thresholds $\theta_\text{th}\in\{\pi/3,\pi/4,\pi/6\}$; lower $\theta_\text{th}$ yields sparser graphs.
For each $\theta_\text{th}$, all methods are evaluated by the NRMSE between their predicted trajectories and the ground truth.

\paragraph{Accuracy.}
\cref{fig:temporal}(a--d) illustrates error dynamics at $\theta_\text{th}=\pi/6$.
While LSODA shows significant deviations from the ground truth, NGS maintains high accuracy.
As quantified in \cref{tab:kuramoto}, NGS consistently achieves the lowest NRMSE across all sparsity levels, outperforming both LSODA and temporal GNN baselines, although its accuracy decreases as $\theta_\text{th}$ decreases.
\cref{fig:temporal}(e) shows the evolution of the mean NRMSE over 50 test systems, with 95\% confidence intervals, up to $T_\text{ext}$.
After an initial transient, the NRMSE exhibits little further growth beyond the training horizon $t \le 5$, in contrast to the linear growth for heat diffusion and the exponential growth for the coupled Rössler system.
Together, these results demonstrate that NGS accurately predicts the ground truth trajectories from sparse, state-dependent temporal graphs and maintains this accuracy beyond the training horizon.

\paragraph{Efficiency.}
The efficiency gap between the numerical solver and deep learning models is most pronounced in this setting.
Across the evaluated thresholds, LSODA's function evaluations and runtime increase as $\theta_\text{th}$ decreases, as shown in \cref{tab:kuramoto}.
In contrast, NGS maintains constant NFE and optimal efficiency due to its architectural simplicity, avoiding the heavy computations of temporal GNN baselines.

\subsection{Empirical Semigroup Consistency} \label{subsec:semigroup}

\begin{table}
    \renewcommand{\arraystretch}{1.2}
    \centering
    \begin{tabular}{@{}cccc@{}}
        \toprule
        System & NRMSE & $(\Delta t_1,\Delta t_2)$ & $R$ \\
        \midrule
        \multirow{3}{*}{\makecell{Heat \\ ($\times 10^{-3}$)}} & \multirow{3}{*}{$4.78 \pm 0.68$} & (1/40, 1/80) & $4.67 \pm 1.06$ \\
        &  & (1/20, 1/40) & $5.05 \pm 0.62$ \\
        &  & (1/20, 1/80) & $6.75 \pm 1.15$ \\
        \midrule
        \multirow{3}{*} {\makecell{Rössler \\ ($\times 10^{-2}$)}} & \multirow{3}{*}{$4.98 \pm 2.92$} & (1/2, 2/3) & $8.32 \pm 3.72$ \\
        &  & (1.0, 2/3) & $4.92 \pm 2.01$ \\
        &  & (1.0, 1/2) & $8.80 \pm 3.56$ \\
        \midrule
        \multirow{3}{*}{\makecell{Kuramoto \\ ($\times 10^{-3}$)}} & \multirow{3}{*}{$1.86 \pm 0.55$} & (1/6, 1/8) & $1.06 \pm 0.17$ \\
        &  & (1/4, 1/6) & $1.35 \pm 0.19$ \\
        &  & (1/4, 1/8) & $1.45 \pm 0.19$ \\
        \bottomrule
    \end{tabular}
    \caption{
        \textbf{Empirical semigroup consistency.}
        NRMSE between NGS predictions and ground truth, together with rollout residuals $R$, is reported as a mean with 95\% confidence interval across 50 test realizations.
        Both metrics are evaluated over $\Ti \cup \Te$, and Kuramoto uses $\theta_\text{th} = \pi/6$.
    } \label{tab:semigroup}
\end{table}

A finite-time transition model should give consistent predictions for different partitions of the same time interval.
For autonomous dynamics, this requirement is expressed by the semigroup property of the exact flow map $\Phi$:
\begin{equation} \label{eq:semigroup}
    \Phi_{\Delta t_1+\Delta t_2}(\bm{S})
    = \Phi_{\Delta t_2}\left(\Phi_{\Delta t_1}(\bm{S})\right).
\end{equation}
We examine this property empirically for the learned NGS map $\widetilde{\Phi}_{\Delta t}(\bm{S}) \equiv \bm{S} + \Delta\widetilde{\bm{S}}_\theta(\bm{S},\Delta t)$.

For each system, we evaluate one trained model on 50 independent test realizations from $\Gi$.
Each pair $(\Delta t_1,\Delta t_2)$ in \cref{tab:semigroup} specifies two fixed-step autoregressive rollouts from the same initial state.
We compute the rollout residual $R$ using the NRMSE formulation, taking the difference between the two predictions at all common timestamps in $\Ti \cup \Te$ and normalizing by the standard deviation of ground truth trajectory.
This residual would vanish under exact semigroup consistency.
The table also reports the NRMSE between NGS predictions and ground truth, recomputed over the same horizon using the non-uniform time intervals employed in \cref{tab:heat,tab:rossler,tab:kuramoto}.

The NRMSE provide a reference sacle for interpreting the sensitivity to time partitioning measured by $R$.
Across the evaluated interval pairs, $R$ is comparable to the NRMSE for heat diffusion and the coupled Rössler system, and smaller for Kuramoto.
These results provide empirical evidence of approximate semigroup consistency across tested dynamical systems, time intervals, and rollout horizons, although no consistency constraint is imposed during training.

\subsection{Spatiotemporal Real-World Datasets}  \label{subsec:real}

\begin{table*}
    \renewcommand{\arraystretch}{1.2}
    \centering
    \begin{tabular}{l*{11}{c}}
        \toprule
        \multirow{2}{*}{Model} & \multicolumn{3}{c}{15 minutes (3 steps)} &      & \multicolumn{3}{c}{30 minutes (6 steps)} &  & \multicolumn{3}{c}{60 minutes (12 steps)}                                       \\
        \cline{2-4} \cline{6-8} \cline{10-12}
        & MAE & RMSE & MAPE &  & MAE & RMSE & MAPE &  & MAE & RMSE & MAPE \\
        \midrule
        HA & 2.88 & 5.59 & 6.80 &  & 2.88 & 5.59 & 6.80 &  & 2.88 & 5.59 & 6.80 \\
        SVR & 1.85 & 3.59 & 3.80 &  & 2.48 & 5.18 & 5.50 &  & 3.28 & 7.08 & 8.00 \\
        ARIMA & 1.62 & 3.30 & 3.50 &  & 2.33 & 4.76 & 5.40 &  & 3.38 & 6.50 & 8.30 \\
        FC-LSTM & 2.05 & 4.19 & 4.80 &  & 2.20 & 4.55 & 5.20 &  & 2.37 & 4.96 & 5.70 \\
        DCRNN & 1.38 & 2.95 & 2.90 &  & 1.74 & 3.97 & 3.90 &  & 2.07 & 4.74 & 4.90 \\
        GWNet & 1.30 & 2.74 & 2.73 &  & 1.63 & 3.70 & 3.67 &  & 1.95 & 4.52 & 4.63 \\
        GMAN & 1.34 & 2.91 & 2.86 &  & 1.63 & 3.76 & 3.68 &  & 1.86 & 4.32 & 4.37 \\
        RGDAN & 1.31 & 2.79 & 2.77 &  & 1.56 & 3.55 & 3.47 &  & 1.82 & 4.20 & 4.28 \\
        AGCRN & 1.37 & 2.87 & 2.94 &  & 1.69 & 3.85 & 3.87 &  & 1.96 & 4.54 & 4.64 \\
        MTGNN & 1.32 & 2.79 & 2.77 &  & 1.65 & 3.74 & 3.69 &  & 1.94 & 4.49 & 4.53 \\
        DGCRN & 1.28 & 2.69 & 2.66 &  & 1.59 & 3.63 & 3.55 &  & 1.89 & 4.42 & 4.43 \\
        STEP$^*$ & 1.26 & 2.73 & 2.59 &  & 1.55 & 3.58 & 3.43 &  & 1.79 & 4.20 & 4.18 \\
        STD-MAE$^*$ & 1.23 & 2.62 & 2.56 &  & 1.53 & 3.53 & 3.42 &  & 1.77 & 4.20 & 4.17 \\
        T-Graphormer & 1.31 & 2.55 & 2.71 &  & 1.52 & 3.14 & 3.23 &  & 1.76 & 3.78 & 3.91 \\
        \midrule
        NGS (ours) & 1.29 & 2.79 & 2.71 &  & 1.60 & 3.70 & 3.61 &  & 1.86 & 4.35 & 4.36 \\
        \midrule
        NGS-IT & 1.32 & 2.87 & 2.79 &  & 1.65 & 3.82 & 3.75 &  & 1.93 & 4.49 & 4.60 \\
        NGS-PM & 1.37 & 2.94 & 2.90 &  & 1.77 & 4.05 & 4.08 &  & 2.09 & 4.85 & 5.01 \\
        \bottomrule
    \end{tabular}
    \caption{
        \textbf{Performance on PEMS-BAY dataset.}
        Comparison of NGS against baselines across 15-, 30-, and 60-minute horizons in the PEMS-BAY traffic dataset.
        We follow the standard split and evaluation protocol and report published numbers under the same protocol.
        When a baseline uses external pretraining, we mark it with an asterisk (*).
        Despite being a versatile simulator without traffic-specific optimizations, NGS achieves competitive accuracy.
        NGS-IT (irregular timing) and NGS-PM (persistent missing) denote NGS trained on data deliberately modified through irregular temporal sampling and persistent node masking, respectively.
    } \label{tab:traffic}
\end{table*}

\begin{table}
    \renewcommand{\arraystretch}{1.2}
    \centering
    \begin{tabular}{lccc}
        \toprule
        \multirow{2}{*}{Models} & 0.5 hours       & 7.5 hours       & 11 hours        \\
                                & (1 Step)        & (15 Steps)      & (22 Steps)      \\
        \midrule
        DCRNN                   & 1.59 $\pm$ 0.17 & 4.10 $\pm$ 0.27 & 4.93 $\pm$ 0.60 \\
        GWNet                   & 1.65 $\pm$ 0.23 & 6.93 $\pm$ 0.58 & 7.93 $\pm$ 0.17 \\
        GGN                     & 1.61 $\pm$ 0.06 & 3.25 $\pm$ 0.04 & 3.04 $\pm$ 0.05 \\
        SGP                     & 1.09 $\pm$ 0.00 & 3.06 $\pm$ 0.11 & 3.13 $\pm$ 0.13 \\
        \midrule
        NGS (ours)              & 1.14 $\pm$ 0.00 & 3.26 $\pm$ 0.04 & 3.27 $\pm$ 0.01 \\
        \bottomrule
    \end{tabular}
    \caption{
        \textbf{Performance on PV-US dataset.}
        Comparison of MAE on the large-scale PV-US dataset.
        NGS demonstrates robust scalability, performing on par with a state-of-the-art model.
        The results for models other than NGS are taken from Table 2 of \citet{cini2023scalable}.
    } \label{tab:pv_us}
\end{table}

\paragraph{Datasets.}
Real-world benchmarks evaluate whether a task-agnostic simulator remains competitive against highly specialized spatiotemporal forecasting models.
We employ two distinct datasets:
\begin{itemize}
    \item \textbf{PEMS-BAY}~\cite{li2018diffusion}: A widely used traffic dataset recording speed at 325 sensors, sampled at 5-minute intervals over 52,116 time steps.
    \item \textbf{PV-US}~\cite{hummon2012sub}: A large-scale energy production dataset covering 5,016 photovoltaic farms with observations aggregated at 30-minute intervals throughout the year 2006.
\end{itemize}
Following standard protocols (\cite{li2018diffusion,hummon2012sub}) of each dataset, we forecast up to 60 minutes (12 steps) for PEMS-BAY and 11 hours (22 steps) for PV-US.

\paragraph{Accuracy.}
\cref{tab:traffic} and \cref{tab:pv_us} summarize the performance of NGS and baseline models in the real-world datasets.
NGS delivers competitive accuracy, matching or exceeding several domain-specific baselines including GMAN and DGCRN.
This is particularly notable given NGS's general-purpose design, which contrasts with the complicated attention and transformer mechanisms employed by specialized spatiotemporal forecasting models.
Detailed experimental configurations and baseline models are provided in Appendix~\ref{app:real}.

We assess NGS under two deliberate modifications to the PEMS-BAY training data while retaining the same model.
NGS-IT (irregular timing) uses irregularly sampled training inputs, and NGS-PM (persistent missing) retains regular time intervals but permanently masks a fixed subset of sensors throughout training.
Both are tested with standard regularly sampled inputs and no artificial sensor masking, following the same evaluation protocol as standard NGS and the baselines.
Detailed preprocessing and training settings are provided in Appendix~\ref{app:real}.

Both training settings yield higher errors than standard NGS, but the degradation remains moderate.
The mean MAPE across the 15-, 30-, and 60-minute horizons increases by approximately $4.3\%$ for NGS-IT and $12.3\%$ for NGS-PM relative to standard NGS.
These results indicate that NGS retains useful predictive accuracy despite irregular or persistently incomplete training observations.
Comparisons with the published baselines should account for these additional training constraints.

Complementary ablation studies on model depth, $\Delta t$ conditioning, and structural gradient masking are presented in Appendix~\ref{app:ablation}.

\section{Conclusion}    \label{sec:conclusion}
In this work, we introduce the Neural Graph Simulator (NGS), a general-purpose framework for learning and simulating complex dynamical systems on graphs directly from noisy and incomplete observations, without requiring explicit governing equations.
NGS estimates $\Delta t$-conditioned transitions with a graph-based local update rule.
Across a diverse benchmark suite, including canonical scientific dynamical systems as well as real-world spatiotemporal datasets, NGS achieves consistently strong accuracy and substantial efficiency gains relative to numerical integration and solver-dependent neural baselines.
Moreover, our structural gradient masking strategy allows NGS to learn faithful dynamics in the presence of missing nodes and observation noise by preventing corrupted supervision from propagating through local neighborhoods.

Several limitations suggest promising directions for future work.
Primarily, NGS currently relies on explicit graph structures.
Extending the framework to systems with unknown topologies by integrating Neural Relational Inference~\cite{kipf2018neural} or differentiable graph structure learning~\cite{franceschi2019learning} would significantly broaden its applicability to domains like evolving social networks.
Moreover, choosing depth $L$ requires careful optimization to balance efficiency with the need to capture long-range dependencies in the dynamical system.
Depth $L$ is a tunable hyperparameter that should reflect the physical propagation scale at the target $\Delta t$.
A larger observation interval warrants a larger $L$, consistent with the finite information-propagation speed underlying our local design.
A potential heuristic is to estimate this signal propagation speed via hop-wise correlation~\cite{cheng2012spatio}.
Lastly, while our gradient masking strategy effectively handles identified missing values, dealing with unidentifiable or stochastic missing patterns remains an open challenge.

Looking forward, the principles underlying NGS extend beyond graph-structured coupled ODEs.
Future iterations could extend the present generalization across graph realizations to transfer across graph families, such as from Erdős–Rényi to Barabási–Albert graphs, or explore adaptation to discretized Euclidean spaces for partial differential equations (PDEs), positioning NGS as a foundational building block for next-generation scientific simulators.

\appendix

\setcounter{equation}{0}
\setcounter{figure}{0}
\setcounter{table}{0}
\renewcommand{\theequation}{A\arabic{equation}}
\renewcommand{\thefigure}{A\arabic{figure}}
\renewcommand{\thetable}{A\arabic{table}}

\section{Implementation Details of NGS}  \label{app:NGS}
This section details the NGS architecture (\cref{fig:schematic}), including specific design choices.
Note that the architectural changes to update ($\phi$) or aggregation functions ($\rho$) may enhance accuracy or efficiency, depending on specific dynamical system requirements.

\paragraph{Inputs.}
NGS processes three primary inputs: current state $\bm{S}(t)$, time step $\Delta t$, and system parameters $C = \{\bm{V}, \bm{E}, \bm{g}\}$ (node, edge, and global coefficients, respectively).

\paragraph{Encoders.}
Input features are mapped into high-dimensional latent vectors $\bm{V}^{(0)}$, $\bm{E}^{(0)}$, and $\bm{g}^{(0)}$ through dedicated encoders.
Encoded $\bm{S}(t)$ and $\Delta t$ are concatenated into node and global features, respectively.
A single node encoder $\text{Enc}^v$ is shared across all node features $\bm{v}^{(0)}_i = \text{Enc}^v(\bm{v}_i),\ {}^\forall i$, ensuring permutation equivariance and efficient structure.
Similarly, an edge encoder $\text{Enc}^e$ is shared across all edge features.
To demonstrate the capability of the NGS in the simplest architecture, all encoders consist of two-layer multi-layer perceptrons (MLPs) with GELU activation.

\paragraph{Graph network.}
Latent vectors are iteratively updated through $L$ Graph Network (GN) layers.
We fix $L=2$ for simplicity, defining the receptive field for node updates.
Each GN layer encompasses three trainable update functions ($\phi_e,\phi_v,\phi_g$), and three permutation-invariant aggregation functions ($\rho_{e\to v}, \rho_{v \to g}, \rho_{e \to g}$).
Update functions are implemented as two-layer MLPs with GELU activation, maintaining the same spirit as the encoders.
The update sequence within a layer is as follows:

First, the edge features are updated in parallel:
\begin{equation}
    \bm{e}_{ij}' = \phi_e \left( \bm{e}_{ij}, \bm{v}_i, \bm{v}_j, \bm{g} \right).
\end{equation}
Next, updated edge features are aggregated into node features.
In the majority of physical systems, the state change of a node is affected by the union of its interactions with all of its neighbors, leading to a summation operation:
\begin{equation}
    \hat{\bm{e}}_i = \rho_{e \to v} \left[ \bm{e}'_{ij}, j \in \mathcal{N}(i) \right] = \sum_{j \in \mathcal{N}(i)}\bm{e}'_{ij}.
\end{equation}
Node features are then updated in parallel using the aggregated edge information:
\begin{equation}
    \bm{v}_i'  = \phi_v \left( \bm{v}_i, \hat{\bm{e}}_i, \bm{g} \right).
\end{equation}
Updated node and edge features are aggregated across the entire graph using a scale- and permutation-invariant operation.
In many dynamical systems, global behavior is governed by the most restrictive local condition, or a bottleneck.
The minimum aggregation can capture such limiting factors more directly than averaging operators.
\begin{equation}
    \bar{\bm{v}} = \rho_{v \to g}[\bm{v}'_i] = \min_i \bm{v}'_i, \quad \bar{\bm{e}} = \rho_{e \to g}[\bm{e}'_{ij}] = \min_{ij} \bm{e}'_{ij}.
\end{equation}
Finally, the global feature is updated using the aggregated node and edge features.
\begin{equation}
    \bm{g}' = \phi_g \left( \bar{\bm{v}}, \bar{\bm{e}}, \bm{g} \right).
\end{equation}

\paragraph{Decoder.}
Following the series of $L$ GN layers, the final node representations $\bm{V}^{(L)}$ are decoded into state increments $\Delta \tilde{\bm{S}}$ via a shared two-layer MLP with GELU activation.
The next state is obtained as $\tilde{\bm{S}}(t + \Delta t) = \bm{S}(t) + \Delta \tilde{\bm{S}}$.

\paragraph{Training details.}
We optimize the model using AdamW with a weight decay of $10^{-2}$ and cosine annealing learning rate scheduling~\cite{loshchilov2017sgdr}.
Exponential moving average (EMA) in parameter space~\cite{izmailov2018averaging} is applied to enhance training stability and generalization.

\section{Experiments on Mathematical Systems}    \label{app:mathematical}
This section provides detailed configurations for the mathematical systems used in the experiments.
For each system, we generate 1000 samples (graphs from $\Gi$ and time intervals from $\Ti$), split into 800 training and 200 validation samples.
Evaluations are conducted on 50 independent samples per domain combination.
\cref{tab:params} summarizes the coefficient $C$ settings used for each dataset.

\begin{table*}
    \renewcommand{\arraystretch}{1.5}
    \setlength{\tabcolsep}{5.7pt}
    \centering
    \begin{tabular}{llcccc}
        \toprule
        & System & State $\bm{S}$ & Node coeff. $\bm{V}$ & Edge coeff. $\bm{E}$ & Global coeff. $\bm{g}$ \\
        \midrule
        \multirow{3}{*}{Mathematical systems} & Heat diffusion  & $T_i~(F_s=1)$ & -- & $d_{ij}~(F_e=1)$ & -- \\
        & Coupled Rössler & $(x_i, y_i, z_i)~(F_s=3)$ & -- & $K_{ij}~(F_e=1)$ & $a, b, c~(F_g=3)$ \\
        & Kuramoto & $(\cos \theta_i, \sin \theta_i)~(F_s=2)$ & $\omega_i~(F_v=1)$ & -- & $K~(F_g=1)$ \\
        \midrule
        \multirow{2}{*}{Real-world datasets}  & PEMS-BAY & \makecell[c]{12-step speeds \\ $(F_s=12)$} & \makecell[c]{Node ID \\ $(F_v=1)$} & $1-W_{ij}~(F_e=1)$ & \makecell[c]{latest TID/DIW \\ $(F_g=2)$}  \\
        & PV-US & \makecell[c]{12-step generation history \\ $(F_s=12)$} & \makecell[c]{Node ID \\ $(F_v=1)$} & $1-W_{ij}~(F_e=1)$ & \makecell[c]{latest TID \\ $(F_g=1)$}  \\
        \bottomrule
    \end{tabular}
    \caption{
        \textbf{Input Instantiations.}
        Specification of state ($\bm{S}$), node ($\bm{V}$), edge ($\bm{E}$), and global ($\bm{g}$) coefficients for each benchmark suite.
        $F$ denotes feature dimension for each input.
        TID and DIW represent \textit{Time In Day} and \textit{Day In Week}, respectively.
    } \label{tab:params}
\end{table*}

\subsection{Heat Diffusion System}  \label{app:heat}
In the heat diffusion system, nodes with temperature $T_i$ interact via edges with dissipation rates $d_{ij}$.
The temperature evolution linearly depends on $d_{ij}$ and the temperature differentials with neighboring nodes.
Initial temperatures are binary ($T_i \in \{0, 1\}$), with the fraction of hot nodes ($T_i=1$) randomly sampled from $[0.2, 0.8]$.
Dissipation rates $d_{ij}$ are drawn from $[0.5, 2.0]$.
We set time domains to $\Ti=[0, 1]$ and $\Te=[1, 20]$ to capture the transient phase before thermal equilibrium, ensuring that the system exhibits rich state evolutions.
Time steps $\Delta t$ are sampled from $[0.01, 0.09]$.

\paragraph{Analytical solution.}
While analytical solutions exist using the graph Laplacian $\mathcal{L}$ (\cref{eq:heat_solution}), the $O(N^3)$ cost of eigenvalue decomposition is prohibitive for large graphs.
Thus, we employ the DOP853 solver for ground truth generation.
\begin{equation}    \label{eq:heat_solution}
    \begin{split}
        \mathcal{L}_{ij} & \equiv \begin{cases}
            \sum_{k \in \mathcal{N}(i)} d_{ik} & \text{if } j = i                    \\
            -d_{ij}                            & \text{if } j \in \mathcal{N}(i)     \\
            0                                  & \text{otherwise},
        \end{cases} \\
        T_i(t) & = \left[\sum_{j=1}^{N} \left( \bm{T}(0) \cdot \bm{\xi}_j \right) e^{-\lambda_j t} \bm{\xi}_j\right]_i,
    \end{split}
\end{equation}
where $\lambda_j$ and $\bm{\xi}_j$ are the $j$th eigenvalue and normalized eigenvector of $\mathcal{L}$, respectively.

\subsection{Coupled Rössler System} \label{app:rossler}
The coupled Rössler system consists of coupled chaotic Rössler attractors in $\mathbb{R}^3$, interacting along the $y$-axis with coupling strength $K_{ij}$.
Initial states are sampled from $(x, y) \in [-4, 4]$, and $z \in [0, 6]$.
The characteristics of the trajectories change drastically depending on the system parameters $a, b$, and $c$.
They are selected as $a, b \in [0.1, 0.3]$ and $c \in [5.0, 7.0]$ to induce chaotic behavior and impose the most challenging simulation scenarios.
Furthermore, coupling constants $K_{ij} \in [0.01, 0.05]$ are kept weak to prevent partial or global synchronization~\cite{omelchenko2012transition, mishra2015chimeralike}, where the positions of a subset or all attractors evolve together.
A Rössler attractor chaotically oscillates in the $xy$-plane, and after an extended period, it suddenly jumps along the $z$-axis.
To accommodate this, time domains are $\Ti=[0, 40]$ and $\Te=[40, 50]$, with $\Delta t \in [0.5, 1.5]$.

Due to the architectural limitations of GDE and NDCN in processing global features, the global system parameters of the coupled Rössler system ($a$, $b$, and $c$) are implemented by assigning identical feature values across all nodes.

\paragraph{Lyapunov exponent.}
We estimate the maximal Lyapunov exponent $\lambda$ to evaluate the precision of simulators.
$\lambda$ is measured as the divergence rate between perturbed trajectories $\bm{s}_i(t)$ and $\bm{s}'_i(t)$:
\begin{equation}    \label{eq:trajectory_discrepancy}
    \delta \bm{S}(t) \equiv \frac{1}{N} \sum_{i=1}^N \|\bm{s}_i(t) - \bm{s}'_i(t)\|_2 , \quad \delta \bm{S}(t) \sim e^{\lambda t} \delta \bm{S}(t_0)
\end{equation}

\subsection{Kuramoto System}    \label{app:kuramoto}
In the Kuramoto system, phase oscillators evolve according to intrinsic angular velocities $\omega_i$ and nonlinear coupling.
Initial phases $\theta_i$ are uniformly sampled within $[-\pi, \pi]$, and $\omega_i$ follows $g(\omega) = \mathcal{N}(0, 1^2)$.
The coupling constant $K \in [0.1, 0.9]$ is chosen below the critical synchronization threshold ($K_c \approx 1.6$) to maintain rich dynamics, as with the coupled Rössler system.
With oscillators rotating at an average period of $2\pi / \langle \vert \omega \vert \rangle = 2\pi / \sqrt{2/\pi} \approx 7.87$, we set $\Ti=[0, 5]$ and $\Delta t \in [0.1, 0.4]$ to capture both multiple and zero rotations in the NGS predictions.
Input of the node state is $(\cos \theta_i, \sin \theta_i)$ to handle periodicity, while the model outputs the phase angle $\theta_i$.

\paragraph{Temporal graph construction.}
For efficient computation, we approximate the dense graph with a sparse graph using a threshold $\theta_\text{th}$.
Edges are considered as active only when $|\sin(\theta_j-\theta_i-\pi/2)| \le \sin(\theta_\text{th})$.
A new graph is dynamically constructed from predicted states $\bm{S}(t)$ at each step during autoregressive rollout.

\paragraph{Baseline models.}
While TG-ODE accommodates non-uniform time steps, it necessitates information about the number of governing equation evaluations performed per time step during data generation using the Euler method.
Given that our data was generated using DOP853, this information was unavailable, requiring us to treat it as a hyperparameter.
As the TDE-GNN implementation assumed non-uniform evaluation times in integer units, we implemented a rescaling procedure using the maximum simulation time to account for the non-uniform floating-point time steps.

\subsection{Computational Details}  \label{app:numerical}
We use \texttt{scipy.integrate} with Numba JIT acceleration for ground truth generation and numerical baseline simulations.
\begin{itemize}
    \item \textbf{Ground truth generation}: DOP853 (explicit Runge--Kutta with order 8).
    \item \textbf{Thresholded Kuramoto simulations}: LSODA (implicit Adams/BDF with stiffness detection).
\end{itemize}
Both solvers adjust step size based on the estimated local error relative to a tolerance of $a + r|s|$, where $s$ represents the solution value.
Numerical solvers run on the CPU in double precision (FP64) with $a=r=10^{-11}$.
Neural inference runs on the GPU in single precision (FP32), with one graph per evaluation.
Runtime benchmarks are performed on an AMD Ryzen Threadripper 9970X CPU and NVIDIA RTX PRO 6000 Blackwell workstation edition GPU.
The complete training process of the NGS model from scratch required less than 24 hours using the aforementioned hardware configuration.

\subsection{Robustness to Data Corruption}    \label{app:corruption}
We analyze the performance of NGS trained under varying corruption levels (noise $\sigma$, missing fraction $p$).

\begin{figure}
    \centering
    \includegraphics[width=\linewidth]{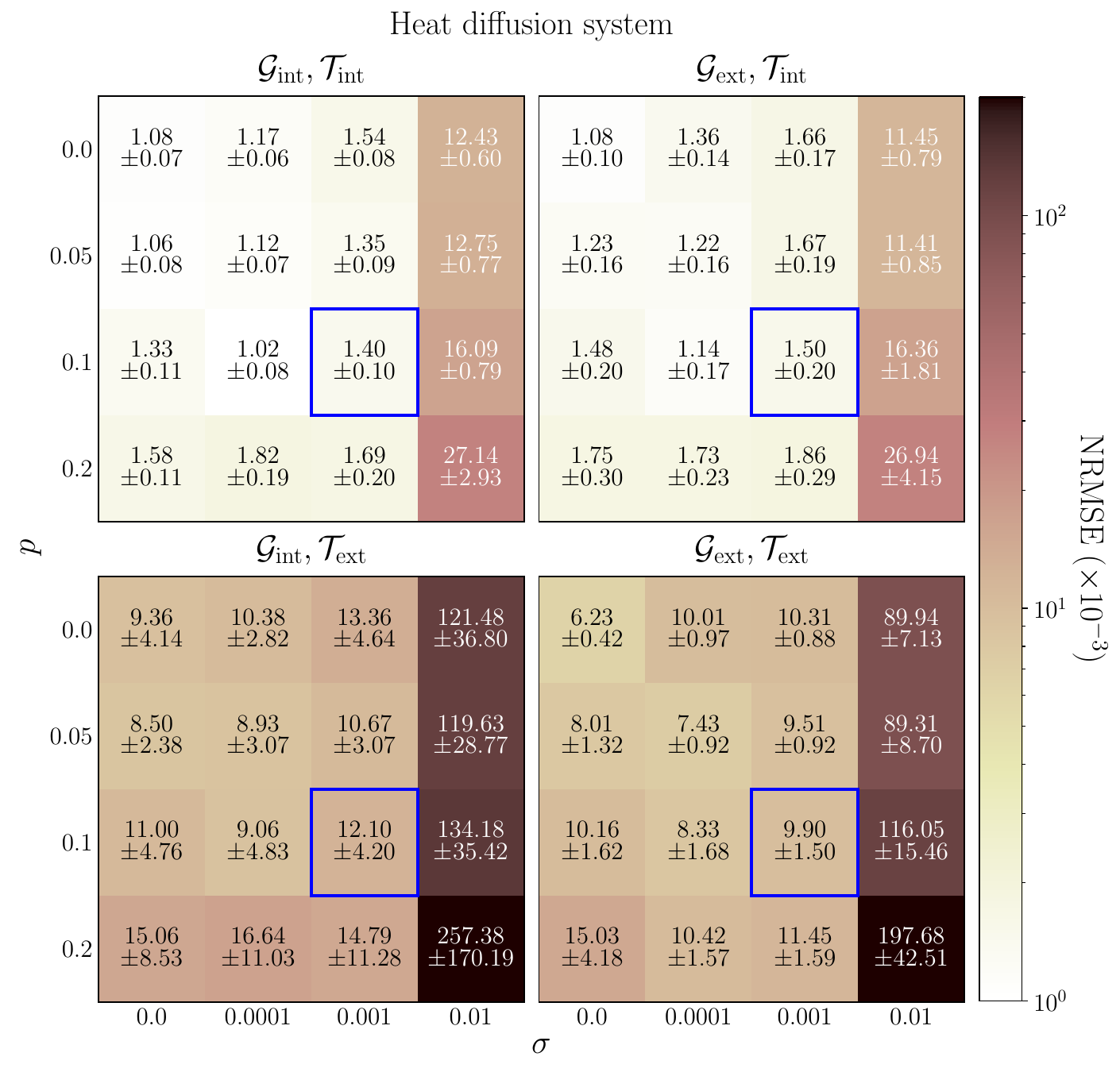}
    \caption{
        \textbf{Heat diffusion system.} Normalized root mean squared error (NRMSE) on a logarithmic scale, in units of $10^{-3}$ vs. $\sigma$ and $p$.
        Blue squares indicate cases with $\sigma=0.001, p=0.1$ discussed in the main text.
    }   \label{fig:heat_result}
\end{figure}

\begin{figure*}
    \centering
    \includegraphics[width=\linewidth]{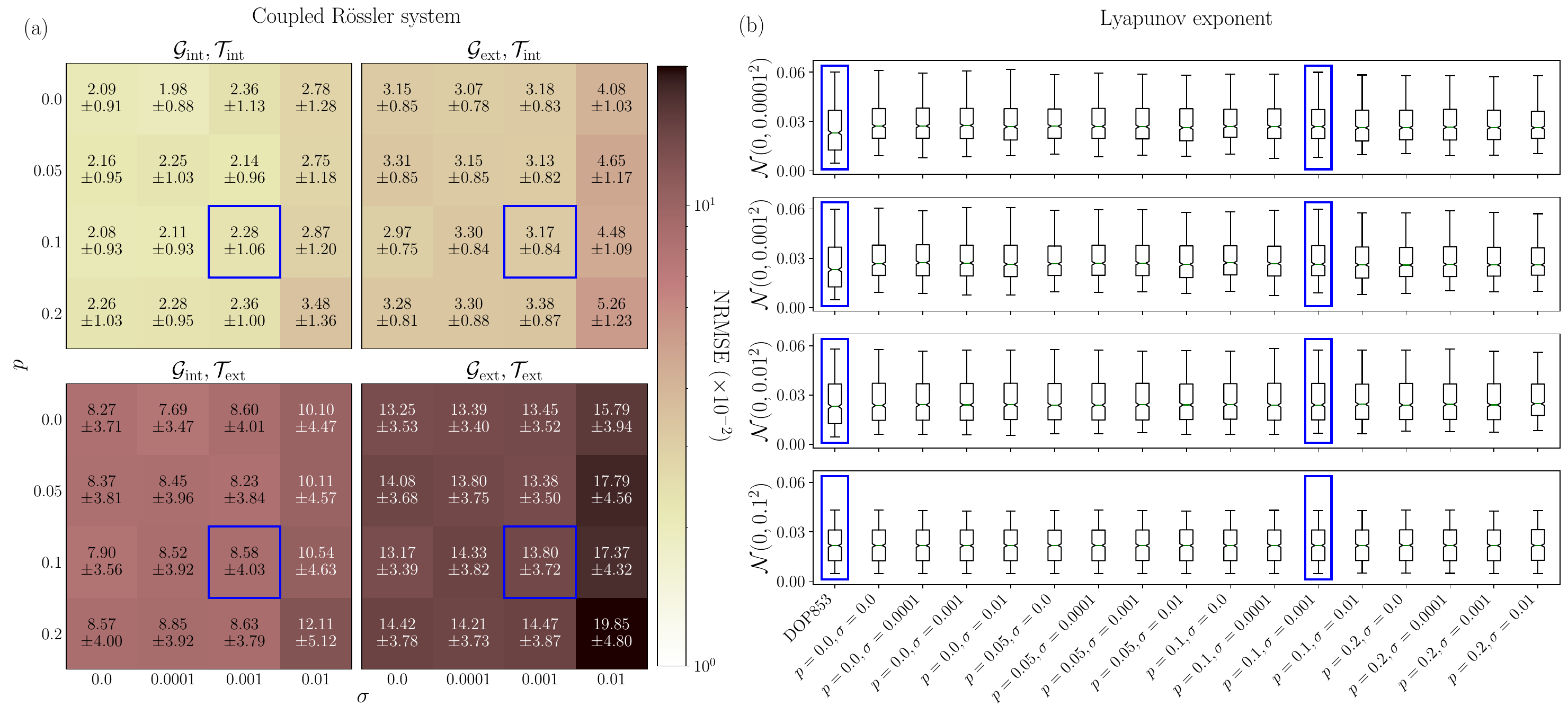}
    \caption{
        \textbf{Coupled Rössler system.}
        NRMSE in units of $10^{-2}$ across various noise levels and missing fractions.
        (b) Lyapunov exponent across varying scales of initial perturbations $\delta \bm{S}(t_0) \sim \mathcal{N}(0, \epsilon^2)$.
    }   \label{fig:rossler_result}
\end{figure*}

NGS exhibits robust performance in the heat diffusion system across diverse noise levels and missing data fractions (\cref{fig:heat_result}) relative to uncorrupted training data ($\sigma=0.0, p=0$).
Notable performance degradation occurs at $\sigma=0.01$, with NRMSE of $(\Gi, \Ti)$ approaching the training data noise level of $\sigma / std(\bm{S}) \approx 40.6 \times 10^{-3}$.

For the coupled Rössler system (\cref{fig:rossler_result}), despite NGS's strong size extrapolation capabilities, its temporal extrapolation performance diminishes due to the system's inherent chaotic behavior.
NGS accurately estimates the Lyapunov exponent even under diverse corruption ranges and initial perturbation scales, confirming it learns the true chaotic attractor rather than overfitting to noise.

\section{Experiments on Real-World Datasets}     \label{app:real}
\subsection{PEMS-BAY and PV-US Datasets}

\paragraph{Datasets.}
We evaluate NGS on two widely used spatiotemporal benchmarks: PEMS-BAY and PV-US.
\begin{itemize}
    \item \textbf{PEMS-BAY}~\cite{li2018diffusion}: Traffic speed data from 325 sensors in the Bay Area (Jan-Jun 2017). Sampled at 5-minute intervals ($T=52,116$).
    \item \textbf{PV-US}~\cite{hummon2012sub}: Solar power generation from 5,016 photovoltaic sites across the US (2006). Aggregated to 30-minute intervals ($T=17,520$) following \citet{cini2023scalable}.
\end{itemize}

\paragraph{Graph construction.}
The adjacency matrices $W_{ij}$ are constructed via a thresholded Gaussian kernel based on distance:
\begin{equation}    \label{eq:traffic_adj}
    W_{ij} = \begin{cases}
        \exp (- d_{ij}^2/s^2) & d_{ij} \le d_c   \\
        0                     & \text{otherwise}
    \end{cases},
\end{equation}
where $d_{ij}$ denotes the directed distance (PEMS-BAY) or geographical distance (PV-US).
$s$ and $d_c$ represent the standard deviation of distances and the threshold distance, respectively, and self-loops are excluded ($W_{ii}=0$).

\paragraph{Data processing.}
Data is split into train, validation, and test sets in chronological order.
PEMS-BAY uses a 7:1:2 ratio, while PV-US uses the first 6 months with a 4:1:1 ratio.
Traffic speeds in PEMS-BAY are normalized using z-scores and solar power generation in PV-US is scaled using min-max scaling.
See \cref{tab:params} for detailed coefficient specifications.

For PEMS-BAY, NGS-IT uses a 12-step input window with time intervals sampled uniformly from $\{5,10,15,20\}$ minutes and shared across all sensors.
These intervals are supplied to the $\Delta t$ encoder as values in $\{1,2,3,4\}$.
NGS-PM retains regular 5-minute inputs while masking a fixed subset of 16 sensors ($p=0.05$) throughout training.
Excluding these sensors and their directed two-hop neighbors ($L=2$) from the training loss leaves 208 trainable nodes.

\subsection{Baseline Models}    \label{app:baseline}
We compare NGS against a comprehensive suite of spatiotemporal forecasting models, categorized as follows.

\paragraph{Temporal-only models.}
\begin{itemize}
    \item Historical Average (HA): This method assumes the traffic flow to be a periodic process and employs weighted averaging of historical data.
    \item Support Vector Regression (SVR): Regression using a support vector machine in a high-dimensional space.
    \item Autoregressive Integrated Moving Average (ARIMA): Classical time series forecasting model combining autoregression and moving averages and addressing non-stationarity to capture trends and seasonality.
    \item FC-LSTM~\cite{sutskever2014sequence}: Encoder-decoder architecture using fully connected layers with LSTM modules.
\end{itemize}

\paragraph{Spatiotemporal neural networks.}
\begin{itemize}
    \item DCRNN~\cite{li2018diffusion}: Models spatial dependency of traffic flow as a diffusion process on directed graphs and captures temporal dependency using a GRU.
    \item GWNet~\cite{wu2019graph}: Graph Wave network, using a self-adaptive adjacency matrix to capture hidden spatial dependencies and stacked dilated 1D convolutions to process long-range temporal sequences.
\end{itemize}

\paragraph{Attention-based models.}
\begin{itemize}
    \item GMAN~\cite{zheng2020gman}: Encoder-decoder architecture with multiple spatiotemporal attention blocks, augmented with gated fusion and a transform attention layer.
    \item RGDAN~\cite{fan2024rgdan}: Integrates graph diffusion attention and temporal attention mechanisms to extract spatiotemporal correlation.
\end{itemize}

\paragraph{Adaptive graph models.}
\begin{itemize}
    \item AGCRN~\cite{bai2020adaptive}: Implements node-specific and spatial dependency adaptive modules, which are combined with a GRU.
    \item MTGNN~\cite{wu2020connecting}: Employs a graph structure derived from the graph learning layer and implements graph and temporal convolutions.
    \item DGCRN~\cite{li2023dynamic}: Employs a hyper-network approach for dynamic graph generation. A graph convolution unit builds a GRU block for spatiotemporal prediction.
\end{itemize}

\paragraph{Pretrained models.}
\begin{itemize}
    \item STEP~\cite{shao2022pre}: Pretrained transformer for recognizing long-term temporal patterns and generating segment-level representations, thereby enhancing the performance of downstream spatiotemporal GNNs.
    \item STD-MAE~\cite{gao2024spatial}: Self-supervised pretrained framework with decoupled spatiotemporal masked autoencoders for capturing clear and complete spatiotemporal heterogeneity.
\end{itemize}

\subsection{Evaluation Metrics} \label{app:evaluation}
Performance is assessed via three standard metrics: mean absolute error (MAE), root mean squared error (RMSE), and mean absolute percentage error (MAPE).
\begin{equation}    \label{eq:traffic_metrics}
    \begin{split}
        \text{MAE}  & = \frac{1}{M} \sum_{m=1}^M \left[ \frac{1}{N} \sum_{i=1}^N \left| \tilde{y}_i(t_m) - y_i(t_m) \right| \right],                                        \\
        \text{RMSE} & = \frac{1}{M} \sum_{m=1}^M  \sqrt{\frac{1}{N} \sum_{i=1}^N \left( \tilde{y}_i(t_m) - y_i(t_m) \right)^2 },                                              \\
        \text{MAPE} & = 100 \times \frac{1}{M} \sum_{m=1}^M \left[ \frac{1}{N} \sum_{i=1}^N \left| \frac{\tilde{y}_i(t_m) - y_i(t_m)}{y_i(t_m)} \right| \right],
    \end{split}
\end{equation}
where $M$ denotes the number of time steps, and $\tilde{y}_i(t_m)$ and $y_i(t_m)$ represent predicted and actual speeds respectively.

\subsection{Visualization of Predictions}    \label{app:traffic_prediction}

\begin{figure*}[ht]
    \centering
    \includegraphics[width=\linewidth]{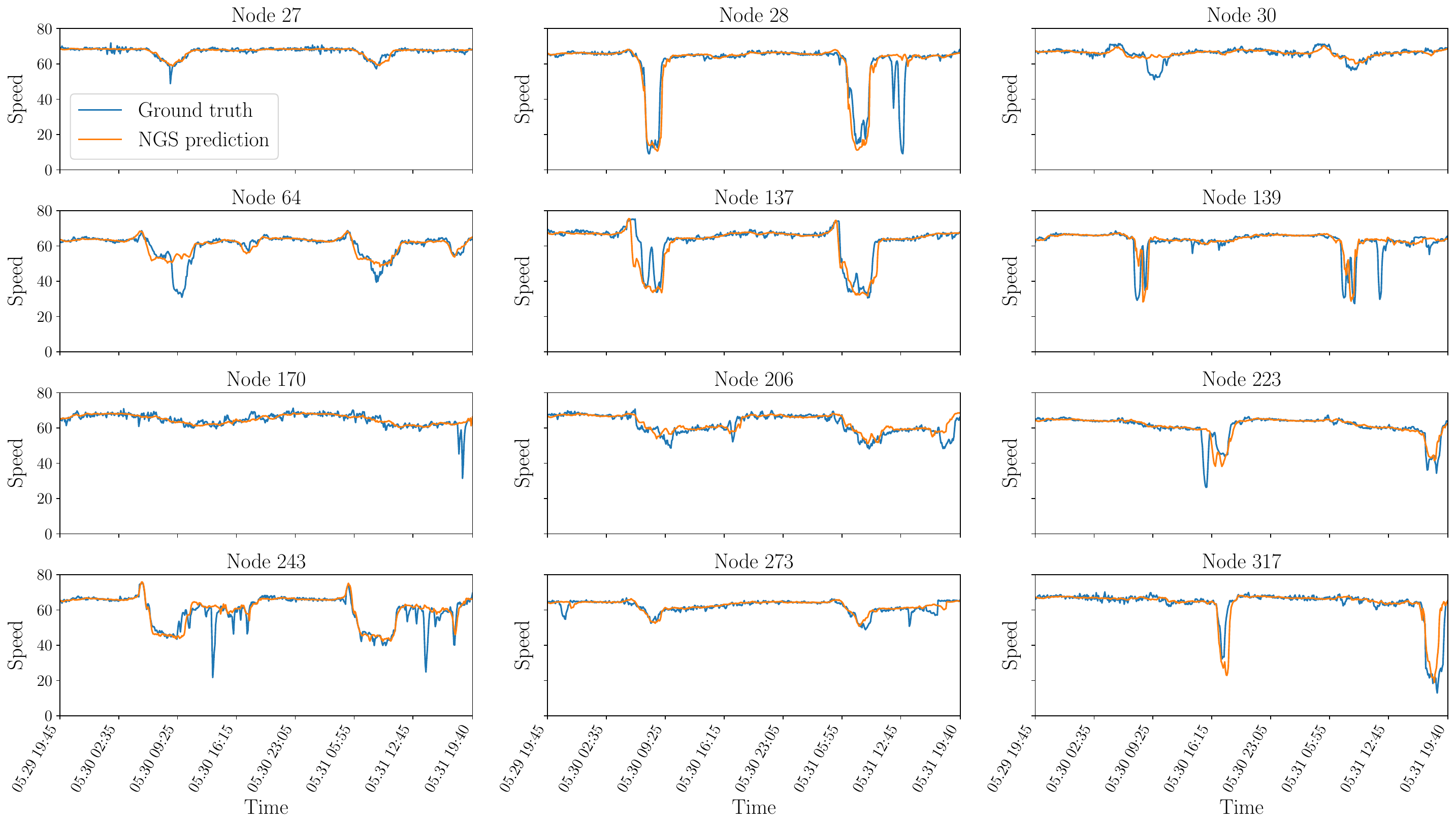}
    \caption{
        \textbf{NGS forecasting of PEMS-BAY.}
        Predictions (orange) and ground truth (blue) for 12 randomly selected sensors.
        NGS accurately captures rush-hour peaks and sudden speed drops, demonstrating robust temporal modeling capabilities.
    }   \label{fig:traffic}
\end{figure*}

\cref{fig:traffic} visualizes NGS predictions on the PEMS-BAY test set.
Despite the highly non-stationary nature of traffic flow, the model effectively tracks sharp transitions.
Full visualizations for all nodes are available in the supplementary material.

\begin{table*}[t]
    \renewcommand{\arraystretch}{1.2}
    \centering
    \begin{tabular}{clccccc}
        \toprule
        System & Configuration & ($\Gi$, $\Ti$) & ($\Gi$, $\Te$) & ($\Ge$, $\Ti$) & ($\Ge$, $\Te$) & Rel. Runtime                                                          \\
        \midrule
        \multirow{9}{*}{\makecell{Heat \\ ($\times 10^{-3}$)}} & NGS (Default) & $1.4 \pm 0.1$ & $12.1 \pm 4.2$ & $1.5 \pm 0.2$ & $9.9 \pm 1.5$ & $1.00\times$ \\
        \cline{2-7}
        & Depth $L$ \\
        & \quad $L=1$ & $9.9 \pm 1.9$ & $88.2 \pm 20.1$ & $10.7 \pm 1.6$ & $67.6 \pm 13.6$ & $0.62\times$ \\
        & \quad $L=3$ & $2.0 \pm 0.4$ & $24.7 \pm 13.7$ & $2.2 \pm 0.3$ & $18.5 \pm 4.3$ & $1.35\times$ \\
        \cline{2-7}
        & $\Delta t$-conditioning \\
        & \quad Without $\Delta t$ & $63.1 \pm 7.5$ & $211.7 \pm 63.9$ & $64.0 \pm 7.67$ & $264.9 \pm 62.1$ & $1.00\times$ \\
        & \quad Euler update & $10.9 \pm 1.4$ & $28.1 \pm 7.9$ & $10.4 \pm 1.5$ & $42.7 \pm 9.6$ & $1.00\times$ \\
        \cline{2-7}
        & Gradient Masking \\
        & \quad Without masking & $12.4 \pm 1.6$ & $103.6 \pm 20.1$ & $14.7 \pm 2.0$ & $95.3 \pm 19.6$ & $1.00\times$ \\
        \midrule
        \multirow{9}{*}{\makecell{Rössler \\ ($\times 10^{-2}$)}} & NGS (Default) & $2.3 \pm 1.1$ & $8.6 \pm 4.0$ & $3.2 \pm 0.8$ & $13.8 \pm 3.7$ & $1.00\times$ \\
        \cline{2-7}
        & Depth $L$ \\
        & \quad $L=1$ & $4.9 \pm 1.6$ & $15.2 \pm 5.6$ & $6.8 \pm 1.4$ & $23.7 \pm 5.5$ & $0.78 \times$ \\
        & \quad $L=3$ & $2.3 \pm 0.9$ & $8.6 \pm 3.7$ & $3.5 \pm 1.0$ & $14.8 \pm 4.0$ & $1.38 \times$ \\
        \cline{2-7}
        & $\Delta t$-conditioning \\
        & \quad Without $\Delta t$ & $88.1 \pm 3.9$ & $127.0 \pm 18.9$ & $98.0 \pm 3.5$ & $322.8 \pm 62.4$ & $1.00 \times$ \\
        & \quad Euler update & $84.7 \pm 4.8$ & $99.7 \pm 6.6$ & $92.6 \pm 4.4$ & $110.1 \pm 5.9$ & $1.00 \times$ \\
        \cline{2-7}
        & Gradient Masking \\
        & \quad Without masking & $4.6 \pm 1.8$ & $15.3 \pm 6.1$ & $6.7 \pm 1.7$ & $23.8 \pm 5.9$ & $1.00 \times$ \\
        \bottomrule
    \end{tabular}
    \caption {
        \textbf{Ablation study on heat diffusion and coupled Rössler systems.}
        Comparison of NRMSE and relative runtime across different configurations.
        The default NGS model uses $L=2$, $\Delta t$-conditioned prediction, and structural gradient masking.
        Relative runtime is measured against the default model for each system.
    } \label{tab:ablation_synth}
\end{table*}

\begin{table*}[t]
    \renewcommand{\arraystretch}{1.2}
    \centering
    \begin{tabular}{l*{13}{c}}
        \toprule
        \multirow{2}{*}{Configuration} & \multicolumn{4}{c}{15 minutes} & \multicolumn{4}{c}{30 minutes} & \multicolumn{4}{c}{60 minutes} & \multirow{2}{*}{Rel. Runtime} \\
        \cline{2-4} \cline{6-8} \cline{10-12}
        & MAE & RMSE & MAPE &  & MAE & RMSE & MAPE &  & MAE & RMSE & MAPE & \\
        \midrule
        NGS (Default) & 1.29 & 2.79 & 2.71 &  & 1.60 & 3.70 & 3.61 &  & 1.86 & 4.35 & 4.36 &  & 1.00$\times$ \\
        \midrule
        Depth $L$ \\
        \quad $L=1$ & 1.28 & 2.76 & 2.71 &  & 1.61 & 3.72 & 3.65 &  & 1.88 & 4.39 & 4.46 &  & 0.86$\times$ \\
        \quad $L=3$ & 1.29 & 2.76 & 2.71 &  & 1.60 & 3.67 & 3.59 &  & 1.86 & 4.33 & 4.37 &  & 1.15$\times$ \\
        \bottomrule
    \end{tabular}
    \caption {
        \textbf{Ablation study of $L$ on the PEMS-BAY dataset.}
        Evaluation of MAE, RMSE, and MAPE for 15-, 30-, and 60-minute horizons. Relative runtime is measured against the default model.
    } \label{tab:ablation_traffic}
\end{table*}

\section{Ablation Studies} \label{app:ablation}
We conduct extensive ablation studies to validate key architectural decisions and training strategies of NGS: depth $L$, $\Delta t$-conditioned prediction, and structural gradient masking.
We utilize the linear heat diffusion system and chaotic Rössler system as synthetic benchmarks, as they allow us to control the data and systematically introduce observation corruption while evaluating ablations across distinct dynamical regimes.
For the real-world PEMS-BAY dataset, which consists of uniformly sampled data without missing observations, we focus on the impact of depth $L$.
All ablated models are trained from scratch and the results are summarized in \cref{tab:ablation_synth} and \cref{tab:ablation_traffic}, respectively.

\subsection{Impact of Depth $L$} \label{app:ablation_L}
NGS applies shared update functions across graph structures and sizes, while the depth $L$ controls the number of node update steps used to aggregate neighborhood information.
We evaluate the model performance with varying $L \in \{1, 2, 3\}$ while keeping other hyperparameters constant.
On both synthetic systems, $L=1$ yields higher errors than $L=2$, whereas $L=3$ increases runtime without improving accuracy.
On PEMS-BAY, prediction accuracy changes little across the tested depths.
This suggests that capturing local dynamics with $L=2$ strikes a balance between accuracy and computational cost across the evaluated systems.

\subsection{Impact of $\Delta t$-Conditioned Prediction} \label{app:ablation_dt}
To assess explicit $\Delta t$-conditioning, we train two variants: ``Without $\Delta t$'' and ``Euler update''.
The former predicts the state increment without time interval information: $\tilde{\bm{S}}(t+\Delta t) = \bm{S}(t) + \Delta \tilde{S} (\bm{S}(t); C)$.
In the latter variant, the GN likewise receives no time interval input, but its output is multiplied by $\Delta t$: $\tilde{\bm{S}}(t+\Delta t) = \bm{S}(t) + \Delta t \tilde{f} (\bm{S}(t); C)$.
This corresponds to one forward Euler step for the learned vector field, hence the name ``Euler update''.

As shown in \cref{tab:ablation_synth}, the ``Without $\Delta t$'' variant exhibits NRMSEs up to tens of times higher than the default NGS on both synthetic systems, with large errors in long-horizon prediction.
Without time interval information, the GN cannot distinguish transitions spanning different intervals from the same initial state.
The ``Euler update'' variant improves accuracy but remains less accurate than the default NGS and the Euler-based NDCN and GDE baselines (\cref{tab:heat}).
Its lower accuracy relative to these baselines may partly reflect differences in state prediction and training: it learns physical-state updates with a single-step loss, whereas NDCN and GDE learn latent dynamics with trajectory-level losses.

Together, the results on synthetic systems support explicit $\Delta t$-conditioning for learning finite-time state increments.
This allows the GN to learn the dependence on elapsed time beyond the proportional scaling imposed by an Euler update.

\subsection{Impact of Structural Gradient Masking} \label{app:ablation_mask}
To robustly learn dynamics from partially observed states, NGS employs a structural gradient masking strategy.
This mechanism blocks gradients originating from missing nodes and their $L$-hop neighborhoods, preventing invalid gradients from missing data from contaminating the learned transition rule.
We ablate this component by allowing gradients to flow through the neighbors of missing nodes during backpropagation.
Specifically, missing values are filled with zeros, and the loss is computed only on observed nodes.
However, unlike in the full model, the gradients from observed nodes are allowed to propagate through the computational graph involving zero-filled missing nodes.
As shown in \cref{tab:ablation_synth}, removing the structural mask leads to higher NRMSEs across all four evaluation domains in both synthetic systems.
These results support the effectiveness of structural gradient masking for learning transitions from partially observed data without relying on explicit imputation.

\section*{Declaration of Competing Interest}
The authors declare that they have no known competing financial interests or personal relationships that could have appeared to influence the work reported in this paper.

\section*{Data and Code Availability}
The fully reproducible code and datasets used in this study are available at \url{https://github.com/hoyunchoi/NGS}.

\section*{Acknowledgements}
This work was supported by the National Research Foundation of Korea (NRF) grant funded by the Korea Government (MSIT) No. RS-2025-00556024 (H. C.), No. RS-2023-00279802 (B. K.), and No. 2022R1A2C1006871 (J. J.), and Korea Institute of Energy Technology Evaluation and Planning (KETEP) and the Ministry of Trade, Industry \& Energy (MOTIE) of the Republic of Korea (No. 20224000000100) (B. K.).
This work is supported by the Center for Advanced Computation at Korea Institute for Advanced Study.

\printcredits

\bibliographystyle{model1-num-names}

\bibliography{reference.bib}

\end{document}